\documentclass[11pt]{article}
\usepackage[letterpaper,margin=1in]{geometry}
\usepackage[T1]{fontenc}
\usepackage{microtype}
\usepackage{xspace}
\usepackage[dvipsnames]{xcolor}
\usepackage{graphicx}
\graphicspath{{CVPR2027/}}
\usepackage{amsmath}
\usepackage{amssymb}
\usepackage{booktabs}
\usepackage{colortbl}
\usepackage[numbers,sort&compress]{natbib}
\usepackage{etoolbox}
\usepackage[format=plain,labelformat=simple,labelsep=period,font=small]{caption}
\usepackage[font=footnotesize,skip=3pt,subrefformat=parens]{subcaption}
\usepackage[hyphens]{url}
\usepackage[shortlabels,inline]{enumitem}

\def\arxivversion{}
\newcommand{\suppref}[1]{\cref{#1}}            
\newcommand{\colfigwidth}{0.6\linewidth}        

\usepackage{amsthm}
\newtheorem{lemma}{Lemma}

\newtheorem{corollary}{Corollary}

\usepackage{dsfont} 
\let\oldmathbb\mathbb
\renewcommand{\mathbb}[1]{%
  \if1#1\mathds{1}\else\oldmathbb{#1}\fi%
}

\usepackage{algorithm}
\usepackage{algpseudocode}
\usepackage{multirow}

\providecommand{\suppref}[1]{supplementary \cref{#1}}
\providecommand{\colfigwidth}{\linewidth}

\definecolor{cvprblue}{rgb}{0.21,0.49,0.74}
\usepackage[breaklinks,colorlinks,allcolors=cvprblue]{hyperref}
\usepackage[capitalize]{cleveref}

\title{JKO-GRPO: Augmenting GRPO with Wasserstein Proximal Steps for Flow-Matching VLAs}

\author{
Jiyong Kwon$^{1,*}$\quad
Yikun Bai$^{2,*,\dagger}$\quad
Amirhossein Mollaali$^{1}$\quad
Guang Lin$^{1,2,\dagger}$\\[6pt]
{\normalsize
$^{1}$School of Mechanical Engineering \quad
$^{2}$Department of Mathematics}\\
{\normalsize
Purdue University, West Lafayette, IN 47907, USA}\\
{\ttfamily\small
\{kwon165,bai195,amollaali,guanglin\}@purdue.edu}
}
\date{}

\begin{document}
\maketitle
{\renewcommand{\thefootnote}{}%
\footnotetext{$^{*}$Equal contribution. \quad $^{\dagger}$Corresponding authors.}}
\begin{abstract}
Flow-based vision-language-action (VLA) models such as $\pi_0,\pi_{0.5}$ generate continuous action chunks by flow matching, and online reinforcement learning (RL) can improve them beyond supervised fine-tuning. Existing methods, such as Flow-GRPO and $\pi_{\mathrm{RL}}$, optimize likelihood ratios of stochastic denoising transitions, which reweight the sampled actions by their advantages but do not use the distances between them. We propose \textbf{JKO-GRPO}, which augments Flow-GRPO with Wasserstein proximal steps, with two contributions:
(1) At each state, we treat the group of sampled actions as an empirical distribution and formulate policy improvement as a Jordan--Kinderlehrer--Otto (JKO) step. We use a soft discrete $c$-transform to construct transport targets from group-relative advantages and pairwise action distances, regress the flow sampler onto these targets using shared noise, and combine this regression with the clipped Flow-GRPO objective.
(2) We extend this approach to sequential VLA control through random-time branching: we select one or multiple action-chunk indices along old-policy trajectories and independently apply group-level JKO updates using candidate continuations from the corresponding states.
\end{abstract}
\section{Introduction}
\label{sec:intro}

Vision-language-action (VLA) models map camera images, language instructions, and proprioceptive states to robot actions, which allows a single policy to perform diverse manipulation tasks from natural-language commands. Flow-based VLAs such as $\pi_0$ and $\pi_{0.5}$~\cite{black2024pi0,intelligence2025pi05} build an action expert on a pretrained vision-language model and generate continuous action chunks by flow matching~\cite{lipman2023flow,liu2023flow}, which supports high-frequency and dexterous control. These models are trained by pretraining on large demonstration datasets followed by supervised fine-tuning (SFT) on the target task. SFT requires costly expert trajectories, its performance is bounded by the quality of these demonstrations, and the resulting policy tends to overfit to them. Reinforcement learning (RL) offers a post-training stage that improves the policy through interaction with the environment, which allows it to exceed the demonstrations and to generalize better to new conditions.

Online RL has recently achieved strong results in generative modeling. Group Relative Policy Optimization (GRPO)~\cite{shao2024deepseekmath} estimates advantages from a group of samples without a value network and has   improved the reasoning ability of large language models~\cite{guo2025deepseek}. Flow-GRPO~\cite{liu2026flow} brings GRPO to flow-matching models by converting the deterministic sampler into a stochastic differential equation, which gives a tractable likelihood for each denoising step, and it improves text-to-image generation in compositional accuracy and text rendering. $\pi_{\mathrm{RL}}$~\cite{pirl2026} extends this approach to flow-based VLAs with learned or converted noise injection, and it raises the success rates of few-shot SFT policies to the level of policies trained on full demonstration datasets. These methods use group-relative advantages to optimize a clipped likelihood-ratio surrogate on the stochastic denoising transitions, where the ratio is taken relative to the rollout policy as in PPO~\cite{schulman2017proximal}. This surrogate raises or lowers the likelihood of each sampled action according to its advantage, and it does not use the distances between the sampled actions.

Optimal transport (OT) provides a complementary geometry, in which an update moves samples. The Jordan--Kinderlehrer--Otto (JKO) scheme~\cite{jordan1998variational} discretizes a Wasserstein gradient flow as a sequence of proximal steps under the Wasserstein distance, and it has been used to train generative models, including input-convex neural networks for large-scale gradient flows~\cite{mokrov2021large,fan2021variational} and normalizing flows trained block by block~\cite{xu2023normalizing}. Wasserstein gradient flows have also been applied to RL: policy optimization can be interpreted as a Wasserstein gradient flow over policies~\cite{zhang2018policy,ziesche2023wasserstein}, and Wasserstein Policy Optimization~\cite{pfau2025wasserstein} projects this flow onto the policy parameters, using the action gradient of a learned action-value function, and SWFP~\cite{sun2025iterative} implements each JKO step of a flow policy against a learned soft Q-function. These methods require a learned critic, whereas a GRPO group provides only scalar rewards at the sampled actions. For robot control, the Wasserstein geometry is natural, since distances between action chunks measure differences in physical motion. In this paper, we augment the clipped likelihood-ratio update of Flow-GRPO with a Wasserstein proximal step and propose JKO-GRPO, a new RL method for flow-based VLA policies. The Wasserstein step treats each denoising trajectory as one sample and acts on the distribution of its endpoint, the action chunk, through distances between the actions in a group; on the denoising transitions, the objective remains that of Flow-GRPO. In this step, a sampled action moves toward another member of its group only when the gain in advantage exceeds the transport cost between them, which gives each sample an action-space direction without a critic.

Our contributions are as follows.
\begin{enumerate}
 \item At each state, we treat the group of actions sampled by the old policy as an empirical distribution and formulate policy improvement as a JKO step of the advantage. Restricted to the group, this step has a closed-form solution given by a discrete $c$-transform: each sampled action moves to the sampled action with the highest advantage net of the transport cost. We soften this discrete update and fit the deterministic flow sampler to the resulting targets by regression through the shared noise, which requires no policy density, and combine this loss with the clipped Flow-GRPO objective, so that the update uses both the likelihood ratios of the denoising transitions and the distances between the sampled actions.
 \item We show that, at the old parameters, the negative gradient of our transport loss is proportional to a masked empirical pathwise update whose action-space direction is constructed from the group without a critic.
\item We extend JKO-GRPO to the $\pi_{\mathrm{RL}}$ setting through random-time branching, which restores the simulator state at selected decision times along each trajectory and applies the update at multiple action chunks. Experiments yield favorable results on ManiSkill, LIBERO, and MetaWorld, with improved success on all 13 unseen ManiSkill scenes.
\end{enumerate}

\section{Related Work}\label{sec:related}

\paragraph{Reinforcement learning for diffusion and flow models.}
Reward fine-tuning of iterative generative models treats the denoising chain as a Markov decision process: DDPO~\cite{black2024ddpo} optimizes the chain with policy gradients, and DPOK~\cite{fan2023dpok} adds KL regularization toward the pre-trained model to limit reward hacking. Deterministic flow-matching samplers lack the per-step likelihoods these methods need; Flow-GRPO~\cite{liu2026flow} obtains them by converting the sampling ODE into a marginal-preserving SDE and applies GRPO~\cite{shao2024deepseekmath}. DGPO~\cite{luo2026dgpo} instead learns from group-level preferences without a policy gradient, which lets it train with an ODE sampler, and Flow-OPD~\cite{fang2026flowopd} distills reward-specific teachers into one student on the student's own rollouts. Other work avoids explicit likelihoods through a surrogate built from the flow-matching loss~\cite{mcallister2026fpo} or a stochastic optimal control formulation~\cite{domingoenrich2024adjoint}. Reward-ranked fine-tuning such as RAFT~\cite{dong2023raft} regresses the policy onto its highest-reward samples. An idealized KL-proximal step reweights sampled actions (\cref{app:bg_fisher_rao}). Our transport term uses pairwise action distances and group-relative advantages to construct regression targets without reward gradients or a critic.

\paragraph{Wasserstein gradient flows in RL and generative modeling.}
Policy optimization has been cast as a Wasserstein gradient flow over policies~\cite{zhang2018policy,ziesche2023wasserstein} and related to the Fokker--Planck equation~\cite{richemond2017wasserstein}. OT-TRPO characterizes continuous-action updates via a one-dimensional dual~\cite{terpin2022trust}, and Wasserstein and Sinkhorn trust-region methods derive explicit updates from advantage and transport costs, primarily for discrete actions~\cite{song2023provably}. Wasserstein natural gradients precondition policy updates~\cite{moskovitz2021efficient}, while practical Wasserstein actor--critic methods use action-value gradients~\cite{zhu2026wasserstein,pfau2025wasserstein}. For flow policies, SWFP~\cite{sun2025iterative} implements each JKO step as a flow block trained against a learned soft Q-function with off-policy replay, and FlowRL~\cite{lv2025flowrl} maximizes a learned Q-function under a Wasserstein-2 constraint toward a behavior-optimal policy defined implicitly from replay.

For generative modeling, the JKO scheme~\cite{jordan1998variational} has been made scalable with input-convex networks~\cite{mokrov2021large,fan2021variational}, normalizing flows~\cite{xu2023normalizing}, and the semi-dual form of unbalanced optimal transport~\cite{choi2024scalable}; ORW-CFM~\cite{fan2025orwcfm} regularizes online reward-weighted flow matching with a Wasserstein distance, concurrent work~\cite{hwang2026reward} fine-tunes one-step generators toward a reward-tilted target with JKO steps and importance weights, and WTF~\cite{mammadov2026wtf} replaces the KL regularizer of reward fine-tuning with a transport cost derived from a pre-trained flow map. Hwang et al.~\cite{hwang2026reward} use an $f$-divergence to a reward-tilted distribution as the JKO energy. Our energy is linear in the group-relative advantage. We use same-state rollout groups to define the discrete subproblem, soften its solution into per-sample targets, and fit the flow sampler via shared-noise regression.

\paragraph{Vision-language-action models and their RL fine-tuning.}
Vision-language-action (VLA) models adapt pre-trained vision-language models to robot control, either by emitting discretized action tokens~\cite{brohan2023rt2,kim2024openvla} or by generating continuous actions with a diffusion or flow action head. $\pi_0$~\cite{black2024pi0} attaches a flow-matching action expert to a vision-language backbone to generate action chunks, and $\pi_{0.5}$~\cite{intelligence2025pi05} adds co-training on heterogeneous data for open-world generalization. Online RL after supervised fine-tuning has improved autoregressive VLAs~\cite{li2025simplevla,zang2025rlinfvla}, but flow-based VLAs inherit the intractable action likelihood of flow matching. $\pi_{\mathrm{RL}}$~\cite{pirl2026}, built on RLinf~\cite{yu2025rlinf}, resolves this with a learnable noise network or an ODE-to-SDE conversion and trains $\pi_0$ and $\pi_{0.5}$ with PPO and GRPO in parallel simulation. We build on its pipeline and add the group-supported $c$-transform step of \cref{sec:ot} to its GRPO objective.

\section{Background}\label{sec:background}

\subsection{Flow-matching policies and Flow-GRPO}\label{sec:bg-flowgrpo}
Let $s\in\mathcal S$ denote an observation, including camera images,
proprioceptive state and the language instruction, and let
$a\in\mathcal A:=\mathbb R^{L\times d}$ denote an action chunk of $L$
low-level actions of dimension $d$, with $\|\cdot\|$ the Frobenius norm.
Let $\mu=\mathcal N(0,I)$ be the base distribution on $\mathcal A$
and $v_\theta:\mathcal A\times[0,1]\times\mathcal S\to\mathcal A$ an
observation-conditioned velocity field. A flow-matching
policy~\cite{lipman2023flow,liu2023flow} draws an action by integrating
\begin{equation}\label{eq:ode}
 \frac{\mathrm d x_u}{\mathrm d u}=v_\theta(x_u,u,s),\qquad x_0=\epsilon\sim\mu,
\end{equation}
from $u=0$ (noise) to $u=1$, and sets $a=f_\theta(\epsilon;s):=x_1$. The policy is
the pushforward
\begin{equation}\label{eq:pushforward}
 \pi_\theta(\cdot\mid s)=\big(f_\theta(\cdot\,;s)\big)_\#\mu .
\end{equation}
Evaluating $\pi_\theta(a\mid s)$ requires integrating the divergence of
$v_\theta$ along \eqref{eq:ode}, which is too costly for policy gradients.

Flow-GRPO~\cite{liu2026flow} replaces \eqref{eq:ode} by an SDE with the
same marginals $p^\theta_u(\cdot\mid s)$ for every $u\in[0,1]$,
\begin{align}
 \mathrm d x_u&=b_\theta(x_u,u,s)\,\mathrm du+\sigma_u\,\mathrm dW_u,\label{eq:sde}\\
 b_\theta(x_u,u,s)&=v_\theta(x_u,u,s)+\tfrac{\sigma_u^2}{2}\nabla_x\log p^\theta_u(x_u\mid s),\nonumber
\end{align}
where $\sigma:[0,1]\to\mathbb R_{\ge0}$ is a noise schedule. Under the linear
interpolation path, the score is an affine function of $x_u$ and
$v_\theta(x_u,u,s)$, so $b_\theta$ is computable from $v_\theta$. On a grid
$0=u_0<\cdots<u_K=1$ with $\Delta_k=u_{k+1}-u_k$, the Euler--Maruyama step
gives the Gaussian transition
\ifdefined\arxivversion
\begin{align}
 p_\theta\big(x^{(k+1)}\mid x^{(k)},s\big)= \mathcal N\!\big(x^{(k+1)};\,x^{(k)}+b_\theta(x^{(k)},u_k,s)\Delta_k,\;\sigma_{u_k}^2\Delta_k I\big),\label{eq:kernel}
\end{align}
\else
\begin{align}
 &p_\theta\big(x^{(k+1)}\mid x^{(k)},s\big)\nonumber\\
 &=\mathcal N\!\big(x^{(k+1)};\,
 x^{(k)}+b_\theta(x^{(k)},u_k,s)\Delta_k,\;\sigma_{u_k}^2\Delta_k I\big),\label{eq:kernel}
\end{align}
\fi
and a denoising trajectory $x^{(0:K)}\in\mathcal A^{K+1}$ with
$x^{(0)}\sim\mu$ and $a=x^{(K)}$.

Let $R:\mathcal A\times\mathcal S\to\mathbb R$ be a reward, where $R(a;s)$
scores a completed action $a$ at observation $s$. The advantage of $a$ under a
policy $\pi$ is
\begin{equation}\label{eq:advfn}
 A^{\pi}(s,a)=R(a;s)-\mathbb E_{a'\sim\pi(\cdot\mid s)}\big[R(a';s)\big].
\end{equation}
GRPO estimates it without a critic: it samples $G$ trajectories
$x^{(0:K)}_1,\ldots,x^{(0:K)}_G$ from $\pi_{\theta_{\mathrm{old}}}$ at the same
$s$ and assigns trajectory $i$ the group-relative advantage
\begin{equation}\label{eq:adv}
 A_i=\frac{R(x^{(K)}_i;s)-\operatorname{mean}_jR(x^{(K)}_j;s)}
 {\operatorname{std}_jR(x^{(K)}_j;s)+\epsilon_0},
\end{equation}
which estimates $A^{\pi_{\theta_{\mathrm{old}}}}(s,x^{(K)}_i)$ up to the
positive factor $1/\operatorname{std}_jR(x^{(K)}_j;s)$; $\epsilon_0>0$ is a
small constant. With the likelihood ratio
$\rho^{(k)}_i(\theta)=p_\theta(x^{(k+1)}_i\mid x^{(k)}_i,s)/
p_{\theta_{\mathrm{old}}}(x^{(k+1)}_i\mid x^{(k)}_i,s)$, Flow-GRPO maximizes
\ifdefined\arxivversion
\begin{align}
 \mathcal J_{\mathrm{FG}}(\theta)=\frac1G\sum_{i=1}^G\frac1K\sum_{k=0}^{K-1}\min\!\Big(\rho^{(k)}_i(\theta)A_i,\;\operatorname{clip}\big(\rho^{(k)}_i(\theta),1-\varepsilon,1+\varepsilon\big)A_i\Big)-\beta\,D_{\mathrm{KL}}\big(\pi_\theta\,\Vert\,\pi_{\mathrm{ref}}\big).\label{eq:flowgrpo}
\end{align}
\else
\begin{align}
 \mathcal J_{\mathrm{FG}}(\theta)
 &=\frac1G\sum_{i=1}^G\frac1K\sum_{k=0}^{K-1}
 \min\!\Big(\rho^{(k)}_i(\theta)A_i,\nonumber\\
 &\qquad\operatorname{clip}\big(\rho^{(k)}_i(\theta),1-\varepsilon,1+\varepsilon\big)A_i\Big)\nonumber\\
 &\quad-\beta\,D_{\mathrm{KL}}\big(\pi_\theta\,\Vert\,\pi_{\mathrm{ref}}\big).\label{eq:flowgrpo}
\end{align}
\fi
The KL term with $\beta>0$ keeps the fine-tuned policy close to the pretrained reference policy $\pi_{\mathrm{ref}}$. In practice, it is replaced by its upper bound, the sum of the per-step KL divergences between the Gaussian transitions \eqref{eq:kernel} of $\pi_\theta$ and $\pi_{\mathrm{ref}}$. Following $\pi_{\mathrm{RL}}$~\cite{pirl2026}, we set $\beta=0$. The reward is attached to the completed action $x^{(K)}_i$, and the update uses
the actions only through likelihood ratios of the transitions
\eqref{eq:kernel}. In sequential decision problems, $\pi_{\mathrm{RL}}$~\cite{pirl2026}
sums \eqref{eq:flowgrpo} over all decisions of an episode.

\subsection{Wasserstein gradient flows and the JKO scheme}\label{sec:bg-jko}
Let $\mathcal P_2(\mathcal A)$ be the set of probability measures on
$\mathcal A$ with finite second moment. For $P,P'\in\mathcal P_2(\mathcal A)$,
\begin{equation}\label{eq:w2}
 W_2^2(P,P')=\min_{\gamma\in\Pi(P,P')}\int_{\mathcal A\times\mathcal A}
 \|x-y\|^2\,\mathrm d\gamma(x,y),
\end{equation}
where $\Pi(P,P')$ is the set of couplings with marginals $P$ and $P'$.
Given $F:\mathcal P_2(\mathcal A)\to\mathbb R$ and a step size $\tau>0$, the
JKO scheme~\cite{jordan1998variational} iterates
\begin{equation}\label{eq:jko}
 P^{n+1}\in\operatorname*{arg\,min}_{P\in\mathcal P_2(\mathcal A)}
 \Big\{F(P)+\frac{1}{2\tau}W_2^2(P,P^n)\Big\}.
\end{equation}
As $\tau\to0$, the iterates converge to the Wasserstein gradient flow
\begin{equation}\label{eq:wgf}
 \partial_t P_t=\nabla\cdot\Big(P_t\,\nabla\frac{\delta F}{\delta P}(P_t)\Big),
\end{equation}
where $\frac{\delta F}{\delta P}$ is the first variation of
$F$~\cite{ambrosio2008gradient}. The JKO step \eqref{eq:jko} is the implicit
Euler discretization of \eqref{eq:wgf}, and $\tau$ controls how far mass
moves in $W_2$ per step. The time $t$ of \eqref{eq:wgf} indexes the
evolution of the action distribution, while the time $u$ of
\eqref{eq:ode}--\eqref{eq:sde} indexes the generation of one sample.

\subsection{Reinforcement learning as a Wasserstein gradient flow}\label{sec:bg-rl-wgf}
For a fixed observation $s$, maximizing expected reward is minimizing
\ifdefined\arxivversion
\begin{align}
 F_s(P)=-\int_{\mathcal A}R(a;s)\,\mathrm dP(a),\qquad\frac{\delta F_s}{\delta P}=-R(\cdot\,;s).\label{eq:negreward}
\end{align}
\else
\begin{align}
 F_s(P)&=-\int_{\mathcal A}R(a;s)\,\mathrm dP(a),\nonumber\\
 \frac{\delta F_s}{\delta P}&=-R(\cdot\,;s).\label{eq:negreward}
\end{align}
\fi
Substituting into \eqref{eq:wgf} gives the continuity equation
$\partial_tP_t+\nabla\cdot\big(P_t\nabla_aR(\cdot\,;s)\big)=0$: each action moves along $\dot a=\nabla_aR(a;s)$. Viewing policy optimization as a Wasserstein gradient flow over action distributions goes back to~\cite{richemond2017wasserstein,zhang2018policy}. The corresponding JKO step from the
current policy $P^{\mathrm{old}}_s=\pi_{\theta_{\mathrm{old}}}(\cdot\mid s)$ is
\begin{align}
 P^{+}_s\in\operatorname*{arg\,min}_{P\in\mathcal P_2(\mathcal A)}
 &\Big\{-\mathbb E_{a\sim P}\big[R(a;s)\big]+\frac{1}{2\tau}W_2^2\big(P,P^{\mathrm{old}}_s\big)\Big\},\label{eq:jko-rl}
\end{align}
which favors higher-reward actions while keeping $P^{+}_s$ within a $W_2$
neighborhood of $P^{\mathrm{old}}_s$.

\section{Method}\label{sec:method}


\begin{figure*}[t]
\centering
\includegraphics[width=0.8\linewidth]{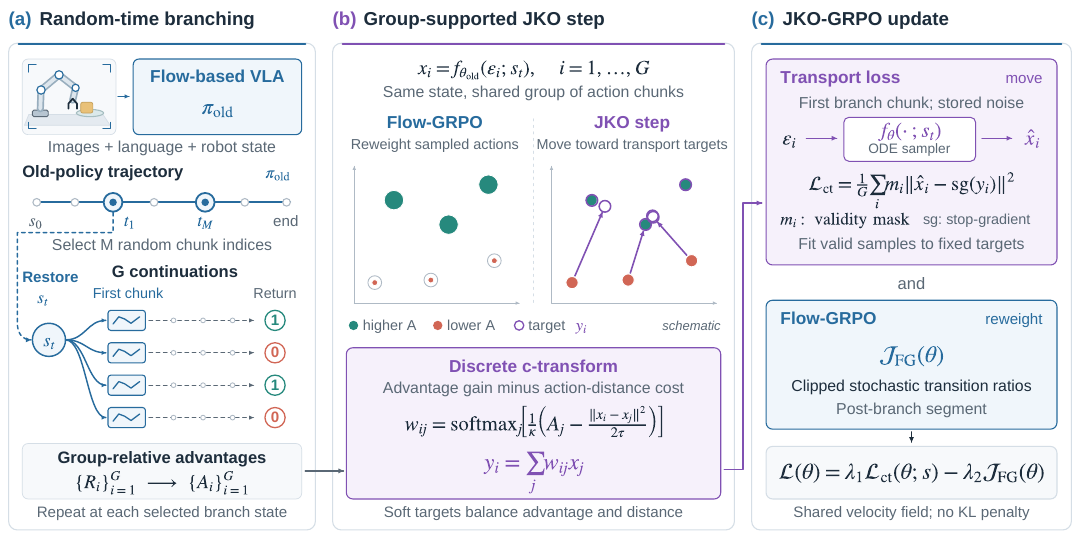}
\caption{\textbf{Overview of JKO-GRPO.} (a) Along each trajectory of the old policy $\pi_{\mathrm{old}}$, we select $M$ random chunk indices, restore the simulator state $s_t$ at each, and form an independent group of $G$ continuations; their returns give group-relative advantages. (b) The first action chunks of a group share the state $s_t$. Flow-GRPO reweights them by advantage. The added JKO step constructs soft $c$-transform targets $y_i$ that trade advantage against action distance. (c) On the first branch chunk, the masked transport loss fits the deterministic ODE sampler to stop-gradient targets, reusing each sample's initial noise $\boldsymbol{\epsilon}_i$. The clipped Flow-GRPO objective acts on the post-branch segment. Both terms train the same velocity field, and the transport loss requires neither a critic nor a policy density.}
\label{fig:overview}
\end{figure*}

\subsection{Group-supported c-transform step}\label{sec:ot}
Flow-GRPO uses the likelihoods of individual stochastic denoising
transitions; our transport term acts on the output of the whole sampler and
uses the distances between the completed actions. Let 
$\boldsymbol{\epsilon}=(\epsilon^{(0)},\ldots,\epsilon^{(K)})\in\mathcal A^{K+1}$ be a noise
sequence with law $\mu_K=\mu^{\otimes(K+1)}$, and let
$f_\theta(\cdot\,;s):\mathcal A^{K+1}\to\mathcal A$ be defined by
\begin{align}
 x^{(0)}&=\epsilon^{(0)},\nonumber\\
 x^{(k+1)}&=x^{(k)}+b_\theta\big(x^{(k)},u_k,s\big)\Delta_k+\sigma_{u_k}\sqrt{\Delta_k}\,\epsilon^{(k+1)},\nonumber\\
 f_\theta(\boldsymbol{\epsilon};s)&=x^{(K)},\label{eq:sampler}
\end{align}
for $k=0,\ldots,K-1$, so that $\pi_\theta(\cdot\mid s)=f_\theta(\cdot\,;s)_\#\mu_K$.
Each step of \eqref{eq:sampler} has the transition \eqref{eq:kernel}. If
$\sigma_{u_k}=0$, then $b_\theta=v_\theta$ and the step is the Euler step of
\eqref{eq:ode}; if $\sigma\equiv0$, $f_\theta$ is the deterministic ODE
sampler, and a hybrid ODE--SDE sampler sets $\sigma_{u_k}>0$ only at some
steps.

\paragraph{JKO step of the advantage.}
Fix an observation $s\in\mathcal S$ and write
$\pi_{\mathrm{old}}=\pi_{\theta_{\mathrm{old}}}$. By \eqref{eq:advfn},
$A^{\pi_{\mathrm{old}}}(s,\cdot)$ differs from $R(\cdot\,;s)$ by a constant,
so replacing $R$ by $A^{\pi_{\mathrm{old}}}(s,\cdot)$ in \eqref{eq:jko-rl}
leaves the minimizer unchanged:
\ifdefined\arxivversion
\begin{align}
 \pi_{\mathrm{new}}(\cdot\mid s)\in\operatorname*{arg\,min}_{P\in\mathcal P_2(\mathcal A)}\Big\{-\int_{\mathcal A}A^{\pi_{\mathrm{old}}}(s,a)\,\mathrm dP(a)+\frac{1}{2\tau}W_2^2\big(P,\pi_{\mathrm{old}}(\cdot\mid s)\big)\Big\}.\label{eq:ct-jko}
\end{align}
\else
\begin{align}
 \pi_{\mathrm{new}}(\cdot\mid s)\in\operatorname*{arg\,min}_{P\in\mathcal P_2(\mathcal A)}
 &\Big\{-\int_{\mathcal A}A^{\pi_{\mathrm{old}}}(s,a)\,\mathrm dP(a)\nonumber\\
 &\quad+\frac{1}{2\tau}W_2^2\big(P,\pi_{\mathrm{old}}(\cdot\mid s)\big)\Big\}.\label{eq:ct-jko}
\end{align}
\fi
Its solution moves each action $a$ to
\ifdefined\arxivversion
\begin{align}
 T_s(a)\in\operatorname*{arg\,max}_{a'\in\mathcal A}\Big[A^{\pi_{\mathrm{old}}}(s,a')-c_\tau(a,a')\Big],\qquad c_\tau(a,a')=\frac{\|a-a'\|^2}{2\tau},\label{eq:ct-map}
\end{align}
\else
\begin{align}
 T_s(a)&\in\operatorname*{arg\,max}_{a'\in\mathcal A}
 \Big[A^{\pi_{\mathrm{old}}}(s,a')-c_\tau(a,a')\Big],\label{eq:ct-map}\\
 c_\tau(a,a')&=\frac{\|a-a'\|^2}{2\tau},\nonumber
\end{align}
\fi
the \textbf{$c_\tau$-transform} of $A^{\pi_{\mathrm{old}}}(s,\cdot)$. Evaluating
\eqref{eq:ct-map} requires $A^{\pi_{\mathrm{old}}}(s,a')$ at every
$a'\in\mathcal A$, whereas the group advantages \eqref{eq:adv} estimate it
only at the sampled actions; evaluating it on all of $\mathcal A$ would
require a learned critic.
We refer to \cref{app:bg_ot} for background on optimal transport and the
$c$-transform, and to \cref{app:method} for all proofs.

\paragraph{Group-supported step.}
GRPO samples $G$ noises $\boldsymbol{\epsilon}_i\sim\mu_K$ at the same observation $s$,
with actions $x_i=f_{\mathrm{old}}(\boldsymbol{\epsilon}_i;s)\in\mathcal A$ and group
advantages $A_i\in\mathbb R$ from \eqref{eq:adv}, $i=1,\ldots,G$. Restricting
\eqref{eq:ct-jko} to measures supported on $\{x_1,\ldots,x_G\}$ and
replacing $\pi_{\mathrm{old}}(\cdot\mid s)$ by the empirical measure
$\frac1G\sum_i\delta_{x_i}$, the restricted problem is
\begin{align}
\min_{P\in\mathcal P(X_G)}\Big\{-\sum_{j=1}^{G}A_j\,P(\{x_j\})
+\frac{1}{2\tau}W_2^2\big(P,\hat\rho\big)\Big\},\label{eq:ct-restricted}
\end{align}
where $X_G=\{x_1,\ldots,x_G\}\subset\mathcal A$, $\mathcal P(X_G)$ is the set of
probability measures on $X_G$, and $\hat\rho=\frac1G\sum_{i=1}^{G}\delta_{x_i}$.
\begin{lemma}[Group-supported JKO step]\label{lem:main_group_jko}
Assume the sampled action locations are distinct. Define $T:X_G\to X_G$, breaking ties in favor of $x_i$ when possible, by
{\ifdefined\arxivversion\else\small\fi
\begin{align}
T(x_i)\in\operatorname*{arg\,max}_{j\in\{1,\ldots,G\}}
\big[A_j-c_\tau(x_i,x_j)\big],\quad i=1,\ldots,G.\label{eq:ct-hard}
\end{align}}
(i) $T_\#\hat\rho$ solves \eqref{eq:ct-restricted}, with optimal value
$\frac1G\sum_{i=1}^{G}\min_{j}\big[c_\tau(x_i,x_j)-A_j\big]$.
(ii) $T(x_i)\neq x_i$ only if some $j$ satisfies $A_j-A_i>c_\tau(x_i,x_j)$;
if all $A_j$ are equal, $T=\mathrm{id}$.
\end{lemma}

See \suppref{app:m_group} for the formal
statement and proof. The map $T$ uses only the rollouts;
$\tau$ sets the exchange rate between advantage gain and transport cost and
is the proximal step size. 

\paragraph{Soft $c$-transform.}
The map $T$ in \eqref{eq:ct-hard} is discontinuous in the advantages: since
each $A_j$ is estimated from $G$ rollouts, a small perturbation of the $A_j$
can switch $T(x_i)$ to another group member. Replacing the $\arg\max$ by a
softmax with temperature $\kappa>0$ gives the targets
\ifdefined\arxivversion
\begin{align}
 w_{ij}=\frac{\exp\big((A_j-c_\tau(x_i,x_j))/\kappa\big)}{\sum_{l=1}^{G}\exp\big((A_l-c_\tau(x_i,x_l))/\kappa\big)},\qquad y_i=\sum_{j=1}^{G}w_{ij}\,x_j\in\mathcal A,\label{eq:ct-target}
\end{align}
\else
\begin{align}
 w_{ij}&=\frac{\exp\big((A_j-c_\tau(x_i,x_j))/\kappa\big)}
 {\sum_{l=1}^{G}\exp\big((A_l-c_\tau(x_i,x_l))/\kappa\big)},\nonumber\\
 y_i&=\sum_{j=1}^{G}w_{ij}\,x_j\in\mathcal A ,\label{eq:ct-target}
\end{align}
\fi
and $y_i\to T(x_i)$ as $\kappa\to0$ when the maximizer is unique
(see \cref{lem:m_soft} in \suppref{app:m_soft}).
For $\kappa>0$, $y_i$ can differ from $x_i$ even when all $A_j$ are equal; such groups carry no reward signal and are skipped by the validity mask $m_i$ (\suppref{app:m_impl}).

\paragraph{Projection onto the sampler.}
The new sampler is coupled to the old one through the shared noise and is
fitted to the targets by
\begin{equation}\label{eq:ct-loss}
 \mathcal L_{\mathrm{ct}}(\theta;s)=\frac1G\sum_{i=1}^{G}m_i\,
 \big\|f_\theta(\boldsymbol{\epsilon}_i;s)-\operatorname{sg}(y_i)\big\|^2 ,
\end{equation}
where $\operatorname{sg}$ is the stop-gradient and $m_i\in\{0,1\}$ is a
validity mask (see \suppref{app:m_impl}). The gradient is back-propagated through
all $K$ steps of \eqref{eq:sampler} and requires neither the density
$\pi_\theta(a\mid s)$ nor a Gaussian transition. If $m_i=1$ for all $i$ and
$\mathcal L_{\mathrm{ct}}=0$, the new policy maps the group noises exactly to
$\{y_i\}$ (\cref{lem:m_proj}); at $\theta=\theta_{\mathrm{old}}$ the negative gradient is $2\tau$ times the masked empirical pathwise update using the group-based surrogate direction $(y_i-x_i)/\tau$.  The targets \eqref{eq:ct-target}
depend only on sampled actions and advantages; the projection \eqref{eq:ct-loss} depends on the flow sampler. In the robot experiments, let $a_i^{\mathrm{roll}}$ denote the chunk executed by the hybrid rollout sampler. Its continuation provides $A_i$. The transport targets and loss use $x_i=f_{\mathrm{old}}^{\mathrm{ODE}}(\epsilon_i^{(0)};s)$ and $f_\theta^{\mathrm{ODE}}(\epsilon_i^{(0)};s)$, respectively. This pairs each hybrid rollout return with the ODE action generated from the same initial noise.


\subsection{Hybrid optimization with Flow-GRPO}\label{sec:hybrid}
We add \eqref{eq:ct-loss} to the clipped Flow-GRPO objective \eqref{eq:flowgrpo} with $\beta=0$ (i.e., without the KL penalty), as in $\pi_{\mathrm{RL}}$~\cite{pirl2026}:
\begin{align}
\mathcal L(\theta)=&\lambda_1\mathcal L_{\mathrm{ct}}(\theta;s)-\lambda_2\mathcal J_{\mathrm{FG}}(\theta)\label{eq:total}
\end{align}
where $\lambda_1,\lambda_2>0$ weight the two terms ($\lambda_2=1$ by default). Each batch from $\pi_{\mathrm{old}}$ is reused for several gradient steps; the clipped ratio $\rho^{(k)}_i(\theta)$ defines the per-transition surrogate for these updates. $\mathcal J_{\mathrm{FG}}$ acts on the transition densities \eqref{eq:kernel} of individual steps; $\mathcal L_{\mathrm{ct}}$ acts on the whole sampler \eqref{eq:sampler} through distances between actions. Both update the same velocity field $v_\theta$. The transport cost limits how far each action moves but not how far the likelihood ratios drift, so we keep the ratio clipping of Flow-GRPO (\cref{sec:exp-abl}).

\Cref{sec:multi} extends \eqref{eq:total} to
multiple action chunks per episode. 

\begin{algorithm}[t]
\caption{JKO-GRPO with a single action chunk}\label{alg:single}
\small
\begin{algorithmic}[1]
\Require velocity field $v_\theta$, group size $G$, step size $\tau$, temperature $\kappa$, weights $\lambda_1, \lambda_2$
\For{each iteration}
  \State $\theta_{\mathrm{old}}\gets\theta$
  \For{each observation $s$ in the batch}
    \For{$i=1,\ldots,G$}
      \State draw $\boldsymbol{\epsilon}_i\sim\mu_K$; compute $x_i=f_{\mathrm{old}}(\boldsymbol{\epsilon}_i;s)$ and its trajectory $x^{(0:K)}_i$ by \eqref{eq:sampler}
      \State evaluate the reward $R(x_i;s)$
    \EndFor
    \State compute $A_1,\ldots,A_G$ by \eqref{eq:adv}
    \State compute the targets $y_1,\ldots,y_G$ by \eqref{eq:ct-target}
  \EndFor
  \For{each gradient step}
    \State compute $\rho^{(k)}_i(\theta)$ by \eqref{eq:kernel} and $\mathcal J_{\mathrm{FG}}(\theta)$ by \eqref{eq:flowgrpo} $(\beta=0)$
    \State compute $\mathcal L_{\mathrm{ct}}(\theta;s)$ by \eqref{eq:ct-loss} for each $s$ in the batch
    \State update $\theta$ by the gradient of \eqref{eq:total} averaged over the batch
  \EndFor
\EndFor
\end{algorithmic}
\end{algorithm}

\subsection{Extension to multiple action chunks}\label{sec:multi}
A VLA policy is queried at every action chunk of an episode, whereas
\cref{sec:ot} compares actions at a single observation $s$. For $M=0$, we form groups at the initial state. At chunk index $h$, the policy observes $s_h\in\mathcal S$, whose camera
images and proprioceptive state are produced by the simulator and whose language instruction is fixed over the
episode. It outputs an action chunk $a_h\in\mathcal A$, and the simulator
executes all $L$ low-level actions of $a_h$ before rendering the next
observation.
An episode of $\pi_{\mathrm{old}}$ is generated by
\ifdefined\arxivversion
\begin{align}
 a_h=f_{\mathrm{old}}(\boldsymbol{\epsilon}_h;s_h),\quad \boldsymbol{\epsilon}_h\sim\mu_K,\qquad s_{h+1}=E(s_h,a_h),\qquad h=0,\ldots,H-1,
\end{align}
\else
\begin{align}
&a_h=f_{\mathrm{old}}(\boldsymbol{\epsilon}_h;s_h),\quad \boldsymbol{\epsilon}_h\sim\mu_K,\qquad
s_{h+1}=E(s_h,a_h)\nonumber\\ &h=0,\ldots,H-1,
\end{align}
\fi
where $s_0$ is the initial observation and $E:\mathcal S\times\mathcal A\to\mathcal S$
is one simulator step, with $s_h$ identified with the full simulator state;
$E$ is accessible only by rollout. The return of an episode is the
task-success indicator $R\in\{0,1\}$ evaluated at its end. 

\Cref{alg:single} requires a group of $G$ actions at a shared observation,
which an episode provides only at $s_0$. Forming such a group at every chunk by branching recursively yields $G^H$ trajectories per episode. We therefore branch at a few random chunks of each episode: random-time branching (\cref{alg:multi}) draws $M$ distinct branch indices $h_1,\ldots,h_M$ uniformly from $\{0,\ldots,H-1\}$.

For each $h_m$, it restores the simulator state reached by
$\pi_{\mathrm{old}}$ at chunk $h_m$ and rolls out $G$ continuations from
$s_{h_m}$ with $\pi_{\mathrm{old}}$ (for one of the $M$ branch points, the source episode itself is one of the $G$);
these $G$ continuations form one group, giving $NM$ groups for $N$ episodes.
Continuation $i$ consists of chunks $h_m,\ldots,H-1$, starts with the branch
chunk $x_i=f_{\mathrm{old}}(\boldsymbol{\epsilon}_i;s_{h_m})$, and has return $R_i$, which replaces $R(x^{(K)}_i;s)$ in \eqref{eq:adv}. All continuations in a group share $s_{h_m}$, so $A_i$ compares action chunks at an identical intermediate state.

In each group, the transport term \eqref{eq:ct-loss} acts on the branch chunks
$x_1,\ldots,x_G$ at $s=s_{h_m}$, and the Flow-GRPO term acts on chunks
$h_m,\ldots,H-1$ of each continuation, summed over chunks as in
$\pi_{\mathrm{RL}}$; the objective averages \eqref{eq:total} over all groups. If $h_m=0$, the group consists of $G$ episodes from a shared initial
state.

\begin{algorithm}[t]
\caption{JKO-GRPO with multiple action chunks}\label{alg:multi}
\small
\begin{algorithmic}[1]
\Require as in \cref{alg:single}; number of chunks $H$, branch points $M$
\For{each iteration}
  \State $\theta_{\mathrm{old}}\gets\theta$; roll out episodes of $\pi_{\mathrm{old}}$ with chunks $a_h=f_{\mathrm{old}}(\boldsymbol{\epsilon}_h;s_h)$ by \eqref{eq:sampler}
  \For{each episode}
    \State draw distinct $h_1,\ldots,h_M$ uniformly from $\{0,\ldots,H-1\}$
    \For{each selected index (or $h_m=0$ if $M=0$)}
      \State restore the simulator state at chunk $h_m$, falling back to $0$ if the source has terminated
      \State roll out $G$ continuations from $s_{h_m}$ with $\pi_{\mathrm{old}}$.
      \State record the branch chunks at $h_m$, $x_i=f_{\mathrm{old}}(\boldsymbol{\epsilon}_i;s_{h_m})$, the trajectories of all chunks by \eqref{eq:sampler}, and the returns $R_i$
      \State compute $A_1,\ldots,A_G$ and the targets $y_1,\ldots,y_G$ as in \cref{alg:single}, with $R_i$ in place of $R(x^{(K)}_i;s)$ and $s=s_{h_m}$
    \EndFor
  \EndFor
  \For{each gradient step}
    \State compute $\mathcal J_{\mathrm{FG}}(\theta)$ and $\mathcal L_{\mathrm{ct}}(\theta;s_{h_m})$ as in \cref{alg:single}, with $\mathcal J_{\mathrm{FG}}$ on all chunks of each continuation and $\mathcal L_{\mathrm{ct}}$ on the branch chunks $x_i$ at $h_1,\ldots,h_M$ of each episode
    \State update $\theta$ as in \cref{alg:single}, averaging over all branch groups
  \EndFor
\EndFor
\end{algorithmic}
\end{algorithm}

\providecommand{\tbd}[1]{\textcolor{red}{#1}}
\providecommand{\suppref}[1]{Supp.~\cref{#1}}
\makeatletter
\@ifundefined{definecolor}{}{\definecolor{defaultgray}{gray}{0.92}}
\providecommand{\dftcell}{\@ifundefined{cellcolor}{}{\cellcolor{defaultgray}}}
\makeatother

\section{Experiments}\label{sec:exp}

\subsection{Controlled 2D study}

\begin{figure}[t]
\centering
\includegraphics[width=\colfigwidth]{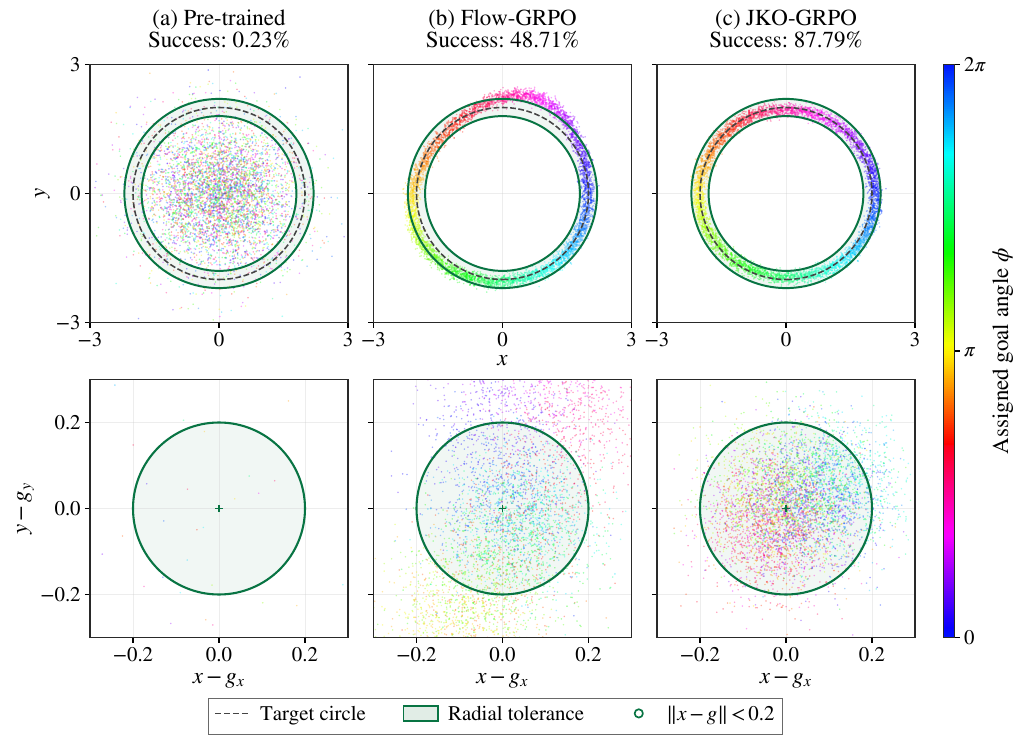}
\caption{\textbf{Conditional 2D task.} Top: outputs before and after RL, colored by goal angle. Bottom: errors $x-g$; the circle marks the success region $\|x-g\|_2<0.2$. One of the nine settings, on 5{,}120 held-out inputs.}
\label{fig:circle-toy}
\end{figure}


We isolate the c-transform term on a goal-conditioned variant
of the 2D study of Hwang et al.~\cite{hwang2026reward}
(\suppref{sec:supp-circle}).
Each group shares a goal $g=2(\cos\phi,\sin\phi)$;
a $10$-step stochastic flow sampler generates $x\in\mathbb{R}^2$
with reward $-\|x-g\|_2^2$ and success criterion
$\|x-g\|_2<0.2$.
Both methods start from the same model pretrained on
$\mathcal{N}(0,I_2)$ and share the sampler, clipped surrogate
without KL penalty, and $16{,}000$ rewarded trajectories.
JKO-GRPO adds the c-transform term with $\lambda_1{=}5$
and $\lambda_2{=}1$, using full SDE replay with stored noise
and no branching. Across nine paired settings, JKO-GRPO raises mean final
held-out success from 72.1\% to 85.8\%, improving success
in seven settings and reward/sample AUC in all nine.
It uses $1.54\times$ the training-loop time and has lower reward/time AUC
(\suppref{tab:supp-circle-results}).
\cref{fig:circle-toy} illustrates one setting;
learning curves are in \suppref{fig:supp-circle-learning-curves}.

\subsection{Robotic VLA setup}\label{sec:exp-setup}

\paragraph{Benchmarks.}
We fine-tune $\pi_{0.5}$~\cite{intelligence2025pi05} from the SFT checkpoints released by $\pi_{\texttt{RL}}$~\cite{pirl2026}.
Our primary benchmark is its ManiSkill pick-and-place suite~\cite{maniskill3}.
We additionally evaluate on the four LIBERO suites~\cite{liu2023libero}, starting from policies fine-tuned on one demonstration per task, and on MetaWorld MT50~\cite{yu2020metaworld}, which uses a single policy for 50 tasks.
For each benchmark, we retain the denoising steps, action-chunk length, and episode horizon of the $\pi_{\texttt{RL}}$ protocol.

\paragraph{Baselines.}
Our main baseline is Flow-GRPO~\cite{liu2026flow}, using the VLA implementation and hybrid ODE--SDE sampler \eqref{eq:sampler} released by $\pi_{\texttt{RL}}$~\cite{pirl2026}.
In this sampler, one randomly chosen denoising step of each action chunk is stochastic, and the other steps are deterministic ODE steps, so the likelihood ratio in \eqref{eq:flowgrpo} has a single term per chunk, at the stochastic step, instead of one term per denoising step.
JKO-GRPO follows the same setting for its likelihood-ratio term.
Flow-GRPO corresponds to $\lambda_1{=}0$, $\lambda_2{=}1$, and $M{=}0$; we examine the contribution of branching in \cref{sec:exp-abl}, and \cref{tab:ablations}a isolates the transport term at $M{=}0$.
We also compare with DGPO~\cite{luo2026dgpo}, which optimizes flow policies without SDE likelihoods.
We follow its deterministic-rollout setting and rescale its temperature to the action-space loss (\suppref{supp:dgpo}).

\paragraph{Training.}
Unless otherwise stated, methods within each benchmark use the same SFT checkpoint, 512 trajectories per iteration, groups of $G{=}8$ continuations from the same simulator state, and the same number of gradient steps.
JKO-GRPO uses $M{=}4$ branch points (\cref{sec:multi}), reusing the source trajectory as a group member within this budget.
We train for 400 iterations on ManiSkill and 40 on LIBERO and MetaWorld.
Robot results are from a single training run per configuration. On LIBERO and MetaWorld, transport hyperparameters were selected by iteration-40 success on the same initial states used for reporting.
Full hyperparameters are in \suppref{tab:supp-hparams}.

\paragraph{Evaluation.}
For the robot benchmarks, we evaluate the deterministic sampler~\eqref{eq:sampler}.
On ManiSkill, we evaluate on 2048 held-out initial layouts every 40 iterations.
On LIBERO and MetaWorld, we evaluate on all 500 initial states per suite every 10 iterations and report the final, iteration-40 checkpoint.
Following the $\pi_{\texttt{RL}}$ protocol, these states are shared between training and evaluation, with no held-out split.
For out-of-distribution evaluation, we follow RL4VLA~\cite{liu2025rl4vla} and evaluate the ManiSkill policies after 400 iterations on 13 unseen scenes spanning Vision, Semantic, and Execution changes, with 512 episodes per scene.

\paragraph{Transport hyperparameters.}
We fix $\lambda_2{=}1$ and $\kappa{=}0.05$ throughout and select benchmark-specific $(\lambda_1,\tau)$ through a small search. The unweighted transport loss ranges from $3.5{\times}10^{-6}$ to $1.3{\times}10^{-3}$ across benchmarks (\suppref{tab:supp-scale}). We use $\lambda_1{=}10$, $\tau{=}1$ on ManiSkill; $\lambda_1{=}1200$ on LIBERO, with $\tau{=}0.3$ on Spatial and $0.1$ on the other suites; and $\lambda_1{=}0.8$, $\tau{=}12$ on MetaWorld. Sensitivity results are reported in \suppref{tab:supp-sens,tab:supp-sens-other}.

\subsection{Main results}\label{sec:exp-main}

\begin{table}[t]
\centering
\small
\setlength{\tabcolsep}{4.5pt}
\begin{tabular}{@{}l|cccc@{}}
\toprule
Iteration & 40 & 80 & 200 & 400 \\
\midrule
SFT & \multicolumn{4}{c}{56.2} \\
Flow-GRPO~\cite{liu2026flow,pirl2026} & 74.5 & 77.1 & 79.5 & 84.7 \\
DGPO~\cite{luo2026dgpo} & 60.5 & 59.0 & 61.3 & 58.1 \\
JKO-GRPO & \textbf{75.6} & \textbf{77.2} & \textbf{89.6} & \textbf{90.3} \\
\bottomrule
\end{tabular}
\caption{\textbf{ManiSkill.} Success (\%) on 2048 held-out layouts. Each RL iteration uses 512 trajectories and 4 gradient steps; JKO-GRPO uses $M{=}4$. Full results, including branch-point ablations, are in \suppref{tab:supp-ms-iters}.}
\label{tab:main}
\end{table}

\begin{table}[t]
\centering
\small
\setlength{\tabcolsep}{2.5pt}
\begin{tabular}{@{}l|ccccc|c@{}}
\toprule
 & \multicolumn{5}{c|}{LIBERO} & MetaWorld \\
 & Spatial & Object & Goal & Long & Avg. & MT50 \\
\midrule
SFT & 82.2 & 96.4 & 86.4 & 45.0 & 77.5 & 47.0 \\
Flow-GRPO & 90.2 & 96.4 & 87.4 & \textbf{61.0} & 83.8 & 61.0 \\
DGPO & 80.8 & 94.8 & 85.6 & 43.8 & 76.3 & 51.4 \\
JKO-GRPO & \textbf{92.0} & \textbf{98.0} & \textbf{87.8} & 60.2 & \textbf{84.5} & \textbf{62.0} \\
\bottomrule
\end{tabular}
\caption{\textbf{LIBERO and MetaWorld.} Success (\%) at iteration 40 on 500 initial states per suite under matched budgets. Avg. shows the unweighted mean across the four LIBERO suites. Full results are in \suppref{tab:supp-iters}.}
\label{tab:other}
\end{table}

\begin{figure*}[t]
\centering
\includegraphics[width=0.7\linewidth]{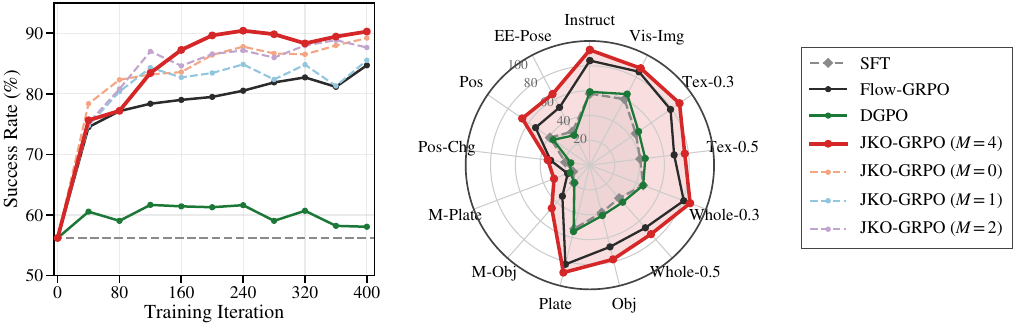}
\caption{\textbf{ManiSkill in-distribution and out-of-distribution evaluation.} Left: success rate on 2048 held-out layouts during training, including the branch-point ablation ($M{=}4$: default). Right: success rate on 13 unseen scenes after 400 iterations, with 512 episodes per scene. All methods use the same trajectory budget per iteration. Complete in-distribution results are in
\suppref{tab:supp-ms-iters}.}
\label{fig:curves}
\end{figure*}

\paragraph{ManiSkill.}
JKO-GRPO reaches 90.3\% success after 400 iterations, exceeding Flow-GRPO by 5.6 percentage points (\cref{tab:main}).
The gap is larger at iteration 200: 89.6\% versus 79.5\%, nearly halving the failure rate from 20.5\% to 10.4\%.
JKO-GRPO reaches Flow-GRPO's final success rate by iteration 160, using 40\% as many iterations under the same per-iteration trajectory budget (\cref{fig:curves}, left).
Under the evaluated configuration, DGPO achieves 58.1\% at iteration 400, compared with 56.2\% for SFT.

\paragraph{LIBERO and MetaWorld.}
JKO-GRPO achieves 84.5\% average success on LIBERO, improving over SFT by 7 points and exceeding Flow-GRPO's 83.8\% (\cref{tab:other}).
It improves over Flow-GRPO on Spatial, Object, and Goal by 1.8, 1.6, and 0.4 points, respectively, while trailing by 0.8 points on Long.
On MetaWorld MT50, it improves success from 61.0\% to 62.0\%.
DGPO remains below SFT on all four LIBERO suites.

\paragraph{Generalization.}
JKO-GRPO outperforms Flow-GRPO on all 13 unseen ManiSkill scenes, with a mean gain of 8.5 points (\cref{fig:curves}, right). The mean gains within the Vision, Semantic, and Execution categories are 6.6, 10.1, and 9.2 points, respectively. Per-scene results are in \suppref{tab:supp-ood}.

\subsection{Analysis}\label{sec:exp-analysis}
\begin{figure}[t]
\centering
\includegraphics[width=\colfigwidth]{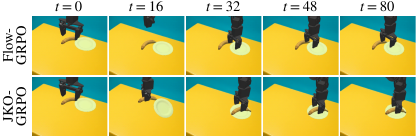}
\caption{\textbf{Rollouts from the same initial state} (ManiSkill, iteration 200). Flow-GRPO reaches the plate without the banana, whereas JKO-GRPO grasps the banana and places it. More rollouts are in \suppref{fig:supp-qual}.}
\label{fig:qual}
\end{figure}

\paragraph{Grasping and placement.}
In a separate failure-logging evaluation of the iteration-200 checkpoints on the same 2048 layouts, success is 80.4\% for Flow-GRPO and 89.6\% for JKO-GRPO.
The fraction of episodes that never hold the object for five consecutive steps decreases from 14.1\% with Flow-GRPO to 5.8\% with JKO-GRPO.
The fraction that grasp but fail to place the object changes less, from 5.6\% to 4.6\%.
The two categories partition all failures; rates use all 2048 episodes as the denominator and are rounded independently.
Most of the observed reduction is in grasp failures
(\cref{fig:qual}).
A finer breakdown is in \suppref{tab:supp-fail}.

\paragraph{Computational overhead.}
For the robot benchmarks, the transport loss requires no additional environment rollouts; its extra computation consists of target construction and a differentiable pass through the ODE sampler. An ODE--SDE replay comparison is provided in \suppref{tab:supp-replay}.
On four H100 GPUs, a ManiSkill iteration takes 261\,s with JKO-GRPO versus 243\,s with Flow-GRPO, an increase of approximately 7\%.
The overhead is 3--9\% on LIBERO and MetaWorld (\suppref{tab:supp-time}).

\subsection{Ablations}\label{sec:exp-abl}

\begin{table}[t]
\centering
\small
\begin{tabular}[t]{l|cc}
\toprule
Regression target & Iter.\ 80 & Iter.\ 160 \\
\midrule
None (Flow-GRPO) & 77.1 & 79.0 \\
RAFT-style target & 79.9 & 81.4 \\
\dftcell Transport target & \dftcell\textbf{82.3} & \dftcell\textbf{83.6} \\
\bottomrule
\end{tabular}
\\[2pt](a) \textbf{Regression target} ($M{=}0$, budget of \cref{tab:main}).
\\[6pt]
\begin{tabular}[t]{c|cc}
\toprule
Ratio clipping & Iter.\ 80 & Iter.\ 200 \\
\midrule
\dftcell On & \dftcell 77.2 & \dftcell\textbf{89.6} \\
Off & \textbf{81.7} & 84.5 \\
\bottomrule
\end{tabular}
\\[2pt](b) \textbf{Ratio clipping} ($M{=}4$, budget of \cref{tab:main}).
\caption{\textbf{Ablations on ManiSkill.} Success rate (\%); default settings are shaded. Both panels use the budget of \cref{tab:main}; (a) uses $M{=}0$ and the transport target with $\lambda_1{=}10$ and $\tau{=}1$, (b) uses $M{=}4$. The branch-point ablation is shown in \cref{fig:curves}.}
\label{tab:ablations}
\end{table}

\paragraph{Branch points.}
We select $M$ distinct chunk indices uniformly along each source trajectory and form an independent continuation group at each restored state (implementation details in \suppref{supp:branch}).
Under the fixed trajectory budget, increasing $M$ trades initial-state coverage for intermediate-state updates: $M{=}0$, 1, 2, and 4 cover 64, 64, 32, and 16 initial states per iteration, respectively.
With $M{=}0$, the transport term alone improves Flow-GRPO by 6.9 points at iteration 200 (\cref{fig:curves}, left).
The $M{=}4$ branching configuration adds another 3.2 points at iteration 200.
Branching learns more slowly early in training but achieves higher success later.

\paragraph{Regression targets.}
A RAFT-style target~\cite{dong2023raft} regresses the first action chunk of each group member toward successful samples from the same group.
RAFT-style regression improves over Flow-GRPO,
while the transport target performs better at both
checkpoints (\cref{tab:ablations}a).
By accounting for both advantage and action-space distance, transport produces source-dependent targets that favor nearby, higher-advantage actions.
RAFT without the GRPO loss and for a smaller trajectory budget are in \suppref{tab:supp-raft}.

\paragraph{Ratio clipping.} Without ratio clipping, JKO-GRPO achieves higher success at iteration 80 (81.7\% vs.\ 77.2\%) but trails the clipped variant by 5.1 percentage points at iteration 200 (\cref{tab:ablations}b). The transport cost penalizes large action displacements, while clipping limits the incentive for large likelihood-ratio changes in the Flow-GRPO surrogate. We therefore keep clipping.

\section{Conclusion}\label{sec:conclusion}
We proposed JKO-GRPO, which augments Flow-GRPO with a Wasserstein proximal step for flow-based VLA policies. Within each GRPO group, the hard JKO subproblem admits a closed-form discrete $c$-transform. We soften this update to construct transport targets and fit the flow sampler to them using shared noise. JKO-GRPO improves success on held-out ManiSkill layouts and all 13 unseen ManiSkill scenes, with smaller average gains on LIBERO and MetaWorld.


\bibliographystyle{CVPR2027/ieeenat_fullname}
\bibliography{ref}

\clearpage
\ifdefined\arxivversion
\onecolumn
\begin{center}
\Large\textbf{Appendix}
\end{center}
\vspace{1em}
\else
\setcounter{page}{1}
\onecolumn
\begin{center}
\Large\textbf{\thetitle}\\[0.5em]
  Supplementary Material
\end{center}
\vspace{1em}
\fi
\appendix


\section{Background: TRPO, PPO and GRPO}
\label{app:background}

\subsection{Discounted return and policy gradient}
\label{app:bg_mdp}

A Markov decision process (MDP) is a tuple
$\mathcal{M} = (\mathcal{S}, \mathcal{A}, P, r, \rho_0, \gamma)$, where
$\mathcal{S}$ is the state space,
$\mathcal{A}$ is the action space,
$P(\cdot \mid s, a) \in \Delta(\mathcal{S})$ is the transition kernel,
$r : \mathcal{S} \times \mathcal{A} \to \mathbb{R}$ is the reward,
$\rho_0 \in \Delta(\mathcal{S})$ is the initial-state distribution, and
$\gamma \in (0,1)$ is the discount factor.
Here $\Delta(\mathcal{X})$ denotes the set of probability distributions on $\mathcal{X}$.
A policy is a map $\pi : \mathcal{S} \to \Delta(\mathcal{A})$, and $\pi(a \mid s)$ denotes its density.
A policy $\pi$ and the MDP generate a trajectory $(s_0, a_0, s_1, a_1, \dots)$ by
\begin{equation}
  s_0 \sim \rho_0, \qquad
  a_t \sim \pi(\cdot \mid s_t), \qquad
  s_{t+1} \sim P(\cdot \mid s_t, a_t), \qquad t = 0, 1, \dots
  \nonumber
\end{equation}
The objective is the expected discounted return
\begin{equation}
  J(\pi) = \mathbb{E}_{\pi}\Big[ \sum_{t=0}^{\infty} \gamma^t \, r(s_t, a_t) \Big].
  \nonumber
\end{equation}
The value function $V^\pi : \mathcal{S} \to \mathbb{R}$, the action-value function $Q^\pi : \mathcal{S} \times \mathcal{A} \to \mathbb{R}$, and the advantage function $A^\pi : \mathcal{S} \times \mathcal{A} \to \mathbb{R}$ are
\begin{align*}
  V^\pi(s) &= \mathbb{E}_{\pi}\Big[ \sum_{t=0}^{\infty} \gamma^t r(s_t, a_t) \,\Big|\, s_0 = s \Big], \qquad A^\pi = Q^\pi - V^\pi ,\\
  Q^\pi(s,a) &= \mathbb{E}_{\pi}\Big[ \sum_{t=0}^{\infty} \gamma^t r(s_t, a_t) \,\Big|\, s_0 = s, a_0 = a \Big].
\end{align*}
The discounted state-visitation distribution $d^\pi \in \Delta(\mathcal{S})$ is
\begin{equation}
  d^\pi(s) = (1-\gamma) \sum_{t=0}^{\infty} \gamma^t \, \Pr(s_t = s \mid \pi).
  \nonumber
\end{equation}

For a parametric policy $\pi_\theta$ with parameters $\theta \in \mathbb{R}^p$, the gradient of $J$ is given by the policy gradient theorem.

\begin{lemma}[Policy gradient~\citep{sutton1999policy}]
\label{lem:pg}
\begin{equation}
  \nabla_\theta J(\pi_\theta)
  = \frac{1}{1-\gamma} \,
  \mathbb{E}_{s \sim d^{\pi_\theta},\, a \sim \pi_\theta(\cdot \mid s)}
  \big[ A^{\pi_\theta}(s,a) \, \nabla_\theta \log \pi_\theta(a \mid s) \big].
  \nonumber
\end{equation}
\end{lemma}
Replacing $Q^{\pi_\theta}$ by $A^{\pi_\theta}$ does not change the gradient, since
$\mathbb{E}_{a \sim \pi_\theta(\cdot \mid s)}[\nabla_\theta \log \pi_\theta(a \mid s)] = 0$ for every $s$; the subtracted term $V^{\pi_\theta}$ acts as a baseline that reduces variance.

The next identity compares the returns of two arbitrary policies and is the starting point of trust-region methods.

\begin{lemma}[Performance difference~\citep{kakade2002approximately}]
\label{lem:pdl}
For any two policies $\pi, \pi'$,
\begin{equation}
  J(\pi') - J(\pi)
  = \frac{1}{1-\gamma} \,
  \mathbb{E}_{s \sim d^{\pi'},\, a \sim \pi'(\cdot \mid s)}
  \big[ A^{\pi}(s,a) \big].
  \nonumber
\end{equation}
\end{lemma}
The right-hand side requires states visited by the new policy $\pi'$, which are not available before $\pi'$ is deployed. TRPO and its successors replace $d^{\pi'}$ by $d^{\pi}$ and control the error of this replacement.

\subsection{Trust region policy optimization (TRPO)}
\label{app:bg_trpo}

Let $\pi_{\mathrm{old}} = \pi_{\theta_{\mathrm{old}}}$ be the current policy, which collects the data.
The importance ratio $w_\theta : \mathcal{S} \times \mathcal{A} \to \mathbb{R}_{>0}$ and the surrogate objective $L_{\mathrm{old}} : \mathbb{R}^p \to \mathbb{R}$ are
\begin{equation}
  w_\theta(s,a) = \frac{\pi_\theta(a \mid s)}{\pi_{\mathrm{old}}(a \mid s)}, \qquad
  L_{\mathrm{old}}(\theta)
  = \mathbb{E}_{s \sim d^{\pi_{\mathrm{old}}},\, a \sim \pi_{\mathrm{old}}(\cdot \mid s)}
  \big[ w_\theta(s,a) \, A^{\pi_{\mathrm{old}}}(s,a) \big].
  \nonumber
\end{equation}
Since $\mathbb{E}_{a \sim \pi_{\mathrm{old}}}[w_\theta A^{\pi_{\mathrm{old}}}] = \mathbb{E}_{a \sim \pi_\theta}[A^{\pi_{\mathrm{old}}}]$, the surrogate $L_{\mathrm{old}}(\theta)/(1-\gamma)$ equals the right-hand side of Lemma~\ref{lem:pdl} with $\pi = \pi_{\mathrm{old}}$, $\pi' = \pi_\theta$, except that the states are drawn from $d^{\pi_{\mathrm{old}}}$ instead of $d^{\pi_\theta}$.
The error of this substitution is bounded by the KL divergence between the two policies.

\begin{lemma}[\citet{schulman2015trust}, Theorem~1]
\label{lem:trpo}
Let $C = 4 \gamma \max_{s,a} |A^{\pi_{\mathrm{old}}}(s,a)| / (1-\gamma)^2$. Then
\begin{equation}
  J(\pi_\theta)
  \;\ge\; J(\pi_{\mathrm{old}})
  + \frac{1}{1-\gamma} L_{\mathrm{old}}(\theta)
  - C \max_{s \in \mathcal{S}} \mathrm{KL}\big( \pi_{\mathrm{old}}(\cdot \mid s) \,\|\, \pi_\theta(\cdot \mid s) \big).
  \nonumber
\end{equation}
\end{lemma}
Maximizing the right-hand side gives monotone improvement of $J$. TRPO replaces the penalty by a constraint on the average KL divergence with radius $\delta > 0$:
\begin{equation}
  \theta_{\mathrm{new}} = \arg\max_{\theta} \; L_{\mathrm{old}}(\theta)
  \quad \text{s.t.} \quad
  \mathbb{E}_{s \sim d^{\pi_{\mathrm{old}}}}
  \big[ \mathrm{KL}\big( \pi_{\mathrm{old}}(\cdot \mid s) \,\|\, \pi_\theta(\cdot \mid s) \big) \big] \le \delta ,
  \nonumber
\end{equation}
and solves it approximately with a linearized objective, a quadratic approximation of the constraint (the Fisher information matrix), conjugate gradient, and a line search.

\subsection{Proximal policy optimization (PPO)}
\label{app:bg_ppo}

PPO~\citep{schulman2017proximal} removes the second-order computation of TRPO and keeps the update close to $\pi_{\mathrm{old}}$ by clipping the importance ratio. With clip range $\eta > 0$, the objective is
\begin{equation}
  L^{\mathrm{clip}}_{\mathrm{old}}(\theta)
  = \mathbb{E}_{s \sim d^{\pi_{\mathrm{old}}},\, a \sim \pi_{\mathrm{old}}(\cdot \mid s)}
  \Big[ \min\big( w_\theta(s,a) \, \hat{A}(s,a), \;
  \mathrm{clip}\big(w_\theta(s,a), 1-\eta, 1+\eta\big) \, \hat{A}(s,a) \big) \Big],
  \nonumber
\end{equation}
where $\mathrm{clip}(w, 1-\eta, 1+\eta) = \min(\max(w, 1-\eta), 1+\eta)$ and $\hat{A}$ is an estimate of $A^{\pi_{\mathrm{old}}}$.
When $\hat{A} > 0$ the objective stops increasing once $w_\theta > 1+\eta$, and when $\hat{A} < 0$ it stops increasing once $w_\theta < 1-\eta$; in both cases the gradient of that sample vanishes, so a single batch cannot move the policy far.
The advantage is estimated with a learned critic $V_\phi : \mathcal{S} \to \mathbb{R}$ and generalized advantage estimation~\citep{schulman2016gae} with parameter $\lambda_{\mathrm{GAE}} \in [0,1]$:
\begin{equation}
  \hat{A}(s_t, a_t)
  = \sum_{l=0}^{\infty} (\gamma \lambda_{\mathrm{GAE}})^l
  \big( r(s_{t+l}, a_{t+l}) + \gamma V_\phi(s_{t+l+1}) - V_\phi(s_{t+l}) \big),
  \nonumber
\end{equation}
and $V_\phi$ is trained by regression onto the empirical returns.

\subsection{Group relative policy optimization (GRPO)}
\label{app:bg_grpo}

GRPO~\citep{shao2024deepseekmath} targets the episodic setting with a single scalar reward at the end of each trajectory, and replaces the critic of PPO by a Monte Carlo baseline computed from a group of trajectories that share the same initial state.
For an initial state $s \sim \rho_0$ (a prompt for a language model, an initial scene for a robot), the old policy generates a group of $G$ trajectories
\begin{equation}
  o_i = \big( (s_{i,1}, a_{i,1}), \dots, (s_{i,T_i}, a_{i,T_i}) \big), \qquad
  s_{i,1} = s, \qquad i = 1, \dots, G,
\end{equation}
where $T_i$ is the length of trajectory $i$, and each trajectory receives a scalar reward $R_i \in \mathbb{R}$.
The group-relative advantage is
\begin{equation}
  A_i = \frac{R_i - \operatorname{mean}(R_1, \dots, R_G)}{\operatorname{std}(R_1, \dots, R_G)}, \qquad i = 1, \dots, G,
\end{equation}
and when all $R_i$ in a group are equal the group carries no learning signal and $A_i = 0$ for all its members.
With the per-step ratio $w_{i,t}(\theta) = \pi_\theta(a_{i,t} \mid s_{i,t}) / \pi_{\mathrm{old}}(a_{i,t} \mid s_{i,t})$, the GRPO objective is
\begin{equation}
  L^{\mathrm{GRPO}}(\theta)
  = \mathbb{E}_{s,\, \{o_i\}_{i=1}^{G} \sim \pi_{\mathrm{old}}}
  \Bigg[ \frac{1}{G} \sum_{i=1}^{G} \frac{1}{T_i} \sum_{t=1}^{T_i}
  \Big( \min\big( w_{i,t}(\theta) A_i, \; \mathrm{clip}(w_{i,t}(\theta), 1-\eta, 1+\eta) A_i \big)
  - \beta \, \hat{D}_{i,t}(\theta) \Big) \Bigg],
\end{equation}
where $\beta \ge 0$ weights a penalty toward a frozen reference policy $\pi_{\mathrm{ref}}$, estimated per step by
\begin{equation}
  \hat{D}_{i,t}(\theta)
  = \frac{\pi_{\mathrm{ref}}(a_{i,t} \mid s_{i,t})}{\pi_\theta(a_{i,t} \mid s_{i,t})}
  - \log \frac{\pi_{\mathrm{ref}}(a_{i,t} \mid s_{i,t})}{\pi_\theta(a_{i,t} \mid s_{i,t})} - 1 ,
  \nonumber
\end{equation}
an unbiased and nonnegative estimator of $\mathrm{KL}(\pi_\theta \,\|\, \pi_{\mathrm{ref}})$ at $s_{i,t}$.
Compared with PPO, GRPO
(i) needs no critic, since the group mean serves as the baseline;
(ii) assigns the same advantage $A_i$ to every step of trajectory $i$; and
(iii) adds the KL term to the loss instead of to the reward.

\subsection{Trust-region updates as a Fisher--Rao gradient flow}
\label{app:bg_fisher_rao}

The analysis in this section concerns the trust-region mechanism of TRPO, which keeps the new policy close to the data-collecting policy $\pi_{\mathrm{old}}$; PPO and GRPO inherit it through ratio clipping. The penalty $\beta\,\mathrm{KL}(\pi_\theta\,\|\,\pi_{\mathrm{ref}})$ in the GRPO objective (\cref{app:bg_grpo}) plays a different role: it is an additional regularizer toward a fixed pretrained policy, used in reward fine-tuning to limit reward hacking and drift from the pretrained model, and it is outside the scope of this section. This corresponds to $\beta = 0$, the setting of $\pi_{\mathrm{RL}}$~\cite{pirl2026} and of all our experiments.

This section shows that the likelihood-ratio updates of TRPO, PPO and GRPO (\cref{app:bg_trpo,app:bg_ppo,app:bg_grpo}) discretize a gradient flow on probability densities under the Fisher--Rao geometry, and \cref{app:bg_flowgrpo_fr} shows that Flow-GRPO is an instance of it.
This flow changes how much probability each action receives and keeps the support of the policy fixed, in contrast with the Wasserstein geometry of the JKO step~\eqref{eq:ct-jko}.
We first define the flow and its proximal step for a general functional, then specify the functional used in policy optimization, and finally recover TRPO, PPO and GRPO.

\paragraph{Setting.}
Fix a state $s \in \mathcal{S}$.
Let $\Omega$ be the space on which the likelihood ratio is evaluated, equipped with a reference measure $\lambda$: for a policy with a tractable action density, $\Omega = \mathcal{A}$ and $\lambda$ is the Lebesgue measure; the denoising chain of Flow-GRPO is treated in \cref{app:bg_flowgrpo_fr}.
All integrals over $\Omega$ are with respect to $\lambda$, and the set of positive probability densities on $\Omega$ is
\begin{align}
\mathcal{D}(\Omega) = \Big\{ P : \Omega \to \mathbb{R}_{>0} \;:\; \int_\Omega P = 1 \Big\}.
\end{align}
Write $\pi_{\mathrm{old}} = \pi_{\mathrm{old}}(\cdot \mid s)$ and $\pi_{\mathrm{ref}} = \pi_{\mathrm{ref}}(\cdot \mid s)$ in $\mathcal{D}(\Omega)$.
For a functional $\mathcal{F} : \mathcal{D}(\Omega) \to \mathbb{R}$, the first variation $\frac{\delta \mathcal{F}}{\delta P}(P) : \Omega \to \mathbb{R}$ is defined up to an additive constant by $\frac{\mathrm{d}}{\mathrm{d}\varepsilon} \mathcal{F}(P + \varepsilon \chi) \big|_{\varepsilon = 0} = \int_\Omega \frac{\delta \mathcal{F}}{\delta P}(P)\, \chi$ for every $\chi : \Omega \to \mathbb{R}$ with $\int_\Omega \chi = 0$.

\paragraph{KL gradient flow.}
The Fisher--Rao metric gives a tangent vector $\chi$ at $P$ (a function with $\int_\Omega \chi = 0$) the inner product
\begin{align}
\langle \chi_1, \chi_2 \rangle_P = \int_\Omega \frac{\chi_1 \chi_2}{P}.
\label{eq:fr_metric}
\end{align}
The Fisher--Rao gradient $\operatorname{grad} \mathcal{F}(P)$ is the tangent vector that represents the first variation in this metric,
\begin{align}
\big\langle \operatorname{grad} \mathcal{F}(P), \chi \big\rangle_P = \int_\Omega \frac{\delta \mathcal{F}}{\delta P}(P)\, \chi \qquad \text{for every } \chi \text{ with } \int_\Omega \chi = 0,
\label{eq:fr_grad_def}
\end{align}
and the gradient flow of $\mathcal{F}$ is $\partial_t \rho_t = -\operatorname{grad} \mathcal{F}(\rho_t)$.

\begin{lemma}[Fisher--Rao gradient]
\label{lem:fr_grad}
$\operatorname{grad} \mathcal{F}(P) = P \big( \frac{\delta \mathcal{F}}{\delta P}(P) - \mathbb{E}_P\big[ \frac{\delta \mathcal{F}}{\delta P}(P) \big] \big)$.
\end{lemma}
\begin{proof}
Let $g = P \big( \frac{\delta \mathcal{F}}{\delta P}(P) - \mathbb{E}_P\big[ \frac{\delta \mathcal{F}}{\delta P}(P) \big] \big)$. Then $\int_\Omega g = 0$, so $g$ is a tangent vector, and for every $\chi$ with $\int_\Omega \chi = 0$,
\begin{align}
\langle g, \chi \rangle_P = \int_\Omega \Big( \frac{\delta \mathcal{F}}{\delta P}(P) - \mathbb{E}_P\Big[ \frac{\delta \mathcal{F}}{\delta P}(P) \Big] \Big) \chi = \int_\Omega \frac{\delta \mathcal{F}}{\delta P}(P)\, \chi,
\end{align}
because the constant integrates to zero against $\chi$. If $h$ is the difference of two tangent vectors satisfying~\eqref{eq:fr_grad_def}, then $\langle h, \chi \rangle_P = 0$ for every tangent $\chi$; taking $\chi = h$ gives $\int_\Omega h^2 / P = 0$, so $h = 0$.
\end{proof}

By \cref{lem:fr_grad}, the gradient flow of $\mathcal{F}$ is
\begin{align}
\partial_t \rho_t = -\rho_t \Big( \frac{\delta \mathcal{F}}{\delta P}(\rho_t) - \mathbb{E}_{\rho_t}\Big[ \frac{\delta \mathcal{F}}{\delta P}(\rho_t) \Big] \Big).
\label{eq:fr_flow}
\end{align}
It changes the density at each point and moves no mass between points, whereas the Wasserstein gradient flow~\eqref{eq:wgf} of the same $\mathcal{F}$ moves mass along $-\nabla \frac{\delta \mathcal{F}}{\delta P}$.
The KL divergence plays for this metric the role that $W_2^2$ plays for the Wasserstein metric: to second order in $\chi$,
\begin{align}
\mathrm{KL}(P + \chi \,\|\, P) = \frac{1}{2} \langle \chi, \chi \rangle_P .
\label{eq:kl_second}
\end{align}
The analogue of the JKO step~\eqref{eq:jko} with step size $\alpha > 0$ is therefore the KL-proximal step
\begin{align}
P^{n+1} \in \operatorname*{arg\,min}_{P \in \mathcal{D}(\Omega)} \Big\{ \mathcal{F}(P) + \frac{1}{\alpha} \mathrm{KL}(P \,\|\, P^n) \Big\},
\label{eq:kl_jko}
\end{align}
which recovers the flow~\eqref{eq:fr_flow} as $\alpha \to 0$.

\paragraph{Implementation of Fisher-Rao Gradient Flow in RL setting.} We now set $\mathcal{F}(P) = -\int_\Omega A \,\mathrm{d}P$, so that $\frac{\delta \mathcal{F}}{\delta P} = -A$ and the Fisher--Rao gradient flow \eqref{eq:fr_flow} becomes
\begin{align}
\partial_t \rho_t = \rho_t \big( A - \mathbb{E}_{\rho_t}[A] \big).
\end{align}
The corresponding step \eqref{eq:kl_jko} is the KL-penalized policy update
\begin{align}
P_{\mathrm{new}} = \operatorname*{arg\,max}_{P \in \mathcal{D}(\Omega)} \Big\{ \mathbb{E}_P[A] - \frac{1}{\alpha} \mathrm{KL}(P \,\|\, \pi_{\mathrm{old}}) \Big\}
= \frac{\pi_{\mathrm{old}}\, e^{\alpha A}}{\mathbb{E}_{\pi_{\mathrm{old}}}[e^{\alpha A}]},
\label{eq:fr_closed}
\end{align}
whose closed form is standard~\citep{peters2010relative,abdolmaleki2018maximum}.
TRPO (\cref{app:bg_trpo}) replaces the penalty by a constraint on the KL divergence and restricts $P$ to the parametric family $\pi_\theta$; PPO and GRPO further replace the constraint by clipping the ratio.

\section{Background: flow-matching policies and Flow-GRPO}
\label{app:background_flow}

\subsection{Flow-matching policies}
\label{app:bg_flow}

A flow-matching policy generates an action by integrating a learned velocity field from Gaussian noise.
Let $\mathcal{A} = \mathbb{R}^{H \times d}$ be the space of action chunks ($H$ future actions of dimension $d$), and let
$v_\theta : \mathcal{A} \times [0,1] \times \mathcal{S} \to \mathcal{A}$ be the velocity field.
For a state $s$, the policy map $f_\theta(\cdot\,; s) : \mathcal{A} \to \mathcal{A}$ sends noise $\epsilon$ to the solution at $u = 1$ of
\begin{equation}
  \frac{\mathrm{d} x^u}{\mathrm{d} u} = v_\theta(x^u, u; s), \qquad x^0 = \epsilon \sim \mathcal{N}(0, I),
  \nonumber
\end{equation}
and the policy is the pushforward of the Gaussian under this map,
\begin{equation}
  \pi_\theta(\cdot \mid s) = f_\theta(\cdot\,; s)_\# \, \mathcal{N}(0, I).
  \nonumber
\end{equation}
The velocity field is trained by flow matching~\citep{lipman2023flow, liu2023flow} on the linear path $x^u = (1-u)\epsilon + u a$ between noise $\epsilon$ and a demonstrated action $a$:
\begin{equation}
  \min_\theta \; \mathbb{E}_{(s,a),\, \epsilon \sim \mathcal{N}(0,I),\, u \sim \mathcal{U}[0,1]}
  \big\| v_\theta(x^u, u; s) - (a - \epsilon) \big\|^2 .
  \nonumber
\end{equation}
The density $\pi_\theta(a \mid s)$ of the pushforward requires integrating the divergence of $v_\theta$ along the ODE, so the importance ratio $w_\theta$ of Sections~\ref{app:bg_trpo}--\ref{app:bg_grpo} is not available in closed form.
Flow-GRPO and Flow-OPD remove this obstacle by replacing the ODE with an SDE whose transitions are Gaussian.

\subsection{Flow-GRPO}
\label{app:bg_flowgrpo}

Flow-GRPO~\citep{liu2026flow} turns the deterministic sampler into a stochastic one with the same marginals and applies GRPO to the resulting denoising transitions.
Let $p_u$ be the marginal density of $x^u$ on the linear path. For this path the score is determined by the velocity,
\begin{equation}
  \nabla_x \log p_u(x) = -\frac{x - u\, v(x,u)}{1-u}, \qquad u \in [0,1),
  \nonumber
\end{equation}
where $v(x,u) = \mathbb{E}[a - \epsilon \mid x^u = x]$ is the exact velocity.

\begin{lemma}[Marginal-preserving SDE]
\label{lem:flow_sde}
For any noise schedule $\sigma : [0,1) \to \mathbb{R}_{\ge 0}$, the SDE
\begin{equation}
  \mathrm{d} x^u = \Big[ v(x^u, u) - \frac{\sigma_u^2}{2(1-u)} \big( x^u - u\, v(x^u, u) \big) \Big] \mathrm{d} u + \sigma_u \, \mathrm{d} W_u ,
  \qquad x^0 \sim \mathcal{N}(0, I),
  \nonumber
\end{equation}
where $W_u$ is a standard Brownian motion on $\mathcal{A}$, has the same marginals $p_u$ as the ODE for every $u$.
\end{lemma}
\begin{proof}
The drift equals $v + \tfrac{\sigma_u^2}{2} \nabla \log p_u$. In the Fokker--Planck equation the added drift contributes $-\nabla \cdot (p_u \tfrac{\sigma_u^2}{2} \nabla \log p_u) = -\tfrac{\sigma_u^2}{2} \Delta p_u$, which cancels the diffusion term $+\tfrac{\sigma_u^2}{2} \Delta p_u$, leaving the continuity equation of the ODE.
\end{proof}

Discretize $[0,1]$ into $0 = u_0 < u_1 < \dots < u_K = 1$ with steps $h_k = u_{k+1} - u_k$, and replace $v$ by $v_\theta$.
One Euler--Maruyama step is a Gaussian transition
\begin{align*}
  \pi_\theta\big( x^{(k+1)} \mid x^{(k)}, s \big) &= \mathcal{N}\big( \mu_\theta(x^{(k)}, u_k; s),\; \sigma_{u_k}^2 h_k \, I \big),\\
  \mu_\theta(x, u; s) &= x + h_k \Big[ v_\theta(x,u;s) - \frac{\sigma_u^2}{2(1-u)} \big( x - u\, v_\theta(x,u;s) \big) \Big],
\end{align*}
whose log-density is available in closed form.
Flow-GRPO uses the schedule $\sigma_u = a_\sigma \sqrt{(1-u)/u}$ with a noise level $a_\sigma > 0$.

The sampler defines a denoising MDP whose state is $(x^{(k)}, u_k, s)$, whose action is $x^{(k+1)}$, and whose only reward is the reward $R$ of the final sample $x^{(K)}$.
For a group of $G$ samples $\{x_i^{(0:K)}\}_{i=1}^{G}$ generated from the same $s$ by $\pi_{\mathrm{old}}$, with advantages $A_i$ as in Section~\ref{app:bg_grpo}, the per-step ratio is
\begin{equation}
  w_{i,k}(\theta) = \frac{\pi_\theta\big( x_i^{(k+1)} \mid x_i^{(k)}, s \big)}{\pi_{\mathrm{old}}\big( x_i^{(k+1)} \mid x_i^{(k)}, s \big)},
  \nonumber
\end{equation}
and the GRPO objective is applied with the average over trajectory steps replaced by the average over denoising steps.
Since the policy and the reference share the covariance $\sigma_{u_k}^2 h_k I$, the KL penalty is exact:
\begin{equation}
  \mathrm{KL}\big( \pi_\theta(\cdot \mid x^{(k)}, s) \,\|\, \pi_{\mathrm{ref}}(\cdot \mid x^{(k)}, s) \big)
  = \frac{\| \mu_\theta - \mu_{\mathrm{ref}} \|^2}{2 \sigma_{u_k}^2 h_k} .
  \nonumber
\end{equation}

\paragraph{Flow-GRPO for VLA policies.}
For a vision-language-action policy, $\pi_{\mathrm{RL}}$~\citep{pirl2026} nests the denoising MDP inside the environment MDP: at every environment step $t$ the policy produces an action chunk by $K$ denoising steps from the observation $s_t$, executes it, and receives the next observation.
In the hybrid ODE--SDE sampler used in this work, one denoising index $k^\star$ per chunk is chosen at random and made stochastic, and all other steps follow the deterministic Euler update.
The log-probability of a chunk is then the log-density of that single Gaussian transition,
\begin{equation}
  \log \pi_\theta(\text{chunk} \mid s_t)
  = \log \mathcal{N}\big( x^{(k^\star+1)};\, \mu_\theta(x^{(k^\star)}, u_{k^\star}; s_t),\; \sigma_{u_{k^\star}}^2 h_{k^\star} I \big),
\end{equation}
which enters the GRPO objective at every environment step, with group members sharing the same initial scene and $R_i$ the binary task success.

\subsection{An idealized Fisher--Rao reading of Flow-GRPO}
\label{app:bg_flowgrpo_fr}

This section omits the reference penalty of Flow-GRPO ($\beta = 0$) and applies to Flow-GRPO the Fisher--Rao reading of the TRPO trust region in \cref{app:bg_fisher_rao}; it is therefore a simplified view of the update.
An idealized distribution-space update for Flow-GRPO can be motivated by the Fisher--Rao gradient flow of \cref{app:bg_fisher_rao}.
Its likelihood ratio is evaluated on the Gaussian transitions of the SDE sampler (\cref{lem:flow_sde}), so $\Omega = \mathcal{A}^{K+1}$ is the space of denoising chains $x^{(0:K)} = (x^{(0)}, \dots, x^{(K)})$, with density
\begin{align}
\pi_\theta^{(0:K)}(x^{(0:K)} \mid s) = \mathcal{N}\big( x^{(0)}; 0, I \big) \prod_{k=0}^{K-1} \mathcal{N}\big( x^{(k+1)};\, \mu_\theta(x^{(k)}, u_k; s),\, \sigma_{u_k}^2 h_k I \big).
\end{align}
For the hybrid sampler of VLA policies, only step $k^\star$ is stochastic, and $\Omega = \mathcal{A}$ is the space of $x^{(k^\star+1)}$ given $x^{(k^\star)}$.
In both cases the advantage depends on $x \in \Omega$ only through the final action $e(x) \in \mathcal{A}$, where $e$ is deterministic: $e(x^{(0:K)}) = x^{(K)}$ for the chain, and $e$ is the composition of the remaining Euler steps for the hybrid sampler.
The KL-proximal step \eqref{eq:kl_jko} on $\Omega$ multiplies $\pi_{\mathrm{old}}$ by $e^{\alpha A(e(x))}$. Since this factor depends only on $e(x)$, the pushforward of the new density under $e$ is the old action density multiplied by $e^{\alpha A}$ with the same normalizer, which is the step \eqref{eq:fr_closed} on $\mathcal{A}$.
This pushforward calculation holds for the idealized KL-proximal step with a fixed endpoint map. The implemented Flow-GRPO objective uses clipped transition ratios and updates a shared parametric sampler.

\paragraph{Remark (Fisher--Rao versus transport).}
Iterating the closed-form update $n$ times from $\rho_0$ gives $\rho_0\, e^{n\alpha A} / \mathbb{E}_{\rho_0}[e^{n\alpha A}]$, which is the solution $\rho_t = \rho_0\, e^{tA} / \mathbb{E}_{\rho_0}[e^{tA}]$ of the flow at $t = n\alpha$; its density ratio satisfies
\begin{align}
e^{-t(\sup A - \inf A)} \le \frac{\rho_t}{\rho_0} = \frac{e^{tA}}{\mathbb{E}_{\rho_0}[e^{tA}]} \le e^{t(\sup A - \inf A)}.
\end{align}
Hence $\rho_t$ has the same support as $\rho_0$, and a region of small probability under $\rho_0$ keeps small probability for bounded $t$.
The Fisher--Rao flow thus changes the weights of the actions that the policy already produces and keeps their locations fixed.
The Wasserstein gradient flow of the same functional $\mathcal{F}(P) = -\mathbb{E}_P[A]$ transports samples instead: $\partial_t \rho_t = -\nabla \cdot (\rho_t \nabla A)$, with particles following $\dot x = \nabla A(x)$, and the JKO step \eqref{eq:ct-jko} is its proximal step.
The two mechanisms combine in the Wasserstein--Fisher--Rao (Hellinger--Kantorovich) gradient flow~\citep{chizat2018unbalanced,liero2018optimal,lu2019accelerating},
\begin{align}
\partial_t \rho_t = -\nabla \cdot (\rho_t \nabla A) + \rho_t \big( A - \mathbb{E}_{\rho_t}[A] \big).
\end{align}

\section{Background: relation between optimal transport (OT) and reinforcement learning}
\label{app:bg_ot}

\subsection{Optimal transport}
\label{app:bg_ot_basic}

Let $\mathcal{X} = \mathbb{R}^n$ and let $\mathcal{P}_2(\mathcal{X})$ be the set of probability measures on $\mathcal{X}$ with finite second moment.
For $\mu, \nu \in \mathcal{P}_2(\mathcal{X})$, the set of couplings is
\begin{equation}
  \Pi(\mu, \nu) = \big\{ \gamma \in \mathcal{P}(\mathcal{X} \times \mathcal{X}) : \gamma(\cdot \times \mathcal{X}) = \mu,\; \gamma(\mathcal{X} \times \cdot) = \nu \big\},
  \nonumber
\end{equation}
and for a map $T : \mathcal{X} \to \mathcal{X}$ the pushforward $T_\# \mu \in \mathcal{P}(\mathcal{X})$ is $T_\# \mu(B) = \mu(T^{-1}(B))$ for every Borel set $B$.
Given a cost $c : \mathcal{X} \times \mathcal{X} \to \mathbb{R}_{\ge 0}$, the Monge problem and its Kantorovich relaxation are
\begin{equation}
  \inf_{T : \, T_\# \mu = \nu} \int c(x, T(x)) \, \mathrm{d}\mu(x),
  \qquad
  \mathrm{OT}_c(\mu, \nu) = \inf_{\gamma \in \Pi(\mu, \nu)} \int c(x, y) \, \mathrm{d}\gamma(x, y).
  \nonumber
\end{equation}
With $c(x,y) = \|x - y\|^2$, the 2-Wasserstein distance is $W_2(\mu, \nu) = \mathrm{OT}_c(\mu, \nu)^{1/2}$, a metric on $\mathcal{P}_2(\mathcal{X})$.
\paragraph{Dual form and c-transform.}
The Kantorovich problem has the dual~\citep{villani2009optimal, santambrogio2015optimal}
\begin{equation}
  \mathrm{OT}_c(\mu, \nu)
  = \sup_{\varphi, \psi} \; \int \varphi \, \mathrm{d}\mu + \int \psi \, \mathrm{d}\nu
  \quad \text{s.t.} \quad \varphi(x) + \psi(y) \le c(x,y) \;\; \forall x, y \in \mathcal{X},
  \nonumber
\end{equation}
over pairs of potentials $\varphi, \psi : \mathcal{X} \to \mathbb{R}$.
For a fixed $\psi$, the objective increases with $\varphi$, and the constraint requires $\varphi(x) \le c(x,y) - \psi(y)$ for every $y$.
The largest admissible $\varphi$ is therefore the c-transform of $\psi$,
\begin{equation}
  \psi^c(x) = \inf_{y \in \mathcal{X}} \big[ c(x, y) - \psi(y) \big],
  \nonumber
\end{equation}
and substituting it gives the semi-dual form
\begin{equation}
  \mathrm{OT}_c(\mu, \nu) = \sup_{\psi} \; \int \psi^c \, \mathrm{d}\mu + \int \psi \, \mathrm{d}\nu .
  \nonumber
\end{equation}
Applying the same argument to $\psi$ with $\varphi = \psi^c$ fixed replaces $\psi$ by $\psi^{cc} \ge \psi$ without decreasing the objective, and $\psi^{ccc} = \psi^c$; hence the optimal potentials can be taken of the form $(\psi^c, \psi)$ with $\psi = \psi^{cc}$ (c-concave).
At an optimal pair and an optimal coupling $\gamma^\star$, the constraint is tight on the support: $\psi^c(x) + \psi(y) = c(x,y)$ for $\gamma^\star$-a.e.\ $(x,y)$, that is, $y$ attains the infimum in $\psi^c(x)$.
For $c(x,y) = \tfrac12 \|x-y\|^2$,
\begin{equation}
  \tfrac12 \|x\|^2 - \psi^c(x) = \big( \tfrac12 \|\cdot\|^2 - \psi \big)^*(x)
  \nonumber
\end{equation}
is a Legendre transform and hence convex, and for absolutely continuous $\mu$ Brenier's theorem~\citep{brenier1991polar} gives a unique optimal coupling $\gamma^\star = (\mathrm{id}, T^\star)_\# \mu$ with
\begin{equation}
  T^\star(x) = \arg\min_{y \in \mathcal{X}} \big[ c(x, y) - \psi(y) \big] \quad \mu\text{-a.e.}
  \nonumber
\end{equation}

\paragraph{Soft c-transform.}
When $\nu = \sum_{j=1}^{m} b_j \delta_{y_j}$ is discrete with weights $b_j > 0$, the infimum is a minimum over the $m$ atoms.
Its entropic smoothing with temperature $\kappa > 0$~\citep{cuturi2013sinkhorn},
\begin{equation}
  \psi^{c,\kappa}(x) = -\kappa \log \sum_{j=1}^{m} b_j \exp\Big( \frac{\psi(y_j) - c(x, y_j)}{\kappa} \Big),
  \nonumber
\end{equation}
converges to $\psi^c(x)$ as $\kappa \to 0$, and its minimizer is replaced by the Gibbs weights
$p_j(x) \propto b_j \exp\big( (\psi(y_j) - c(x, y_j)) / \kappa \big)$ over the atoms, which concentrate on $\arg\min_j [c(x, y_j) - \psi(y_j)]$ as $\kappa \to 0$.

\subsection{Wasserstein gradient flow and the JKO scheme}
\label{app:bg_wgf}

\paragraph{First variation.}
Let $F : \mathcal{P}_2(\mathcal{X}) \to \mathbb{R} \cup \{+\infty\}$ be a functional.
Its first variation at $\rho$ is the function $\frac{\delta F}{\delta \rho}(\rho) : \mathcal{X} \to \mathbb{R}$, defined up to an additive constant, such that
\begin{equation}
  \frac{\mathrm{d}}{\mathrm{d}\varepsilon} F(\rho + \varepsilon \chi) \Big|_{\varepsilon = 0}
  = \int_{\mathcal{X}} \frac{\delta F}{\delta \rho}(\rho)(x) \, \mathrm{d}\chi(x)
  \nonumber
\end{equation}
for every signed measure $\chi$ with $\int \mathrm{d}\chi = 0$ and $\rho + \varepsilon \chi \in \mathcal{P}_2(\mathcal{X})$ for small $\varepsilon > 0$.
Two examples are used below:
\begin{equation}
  F(\rho) = \int U \, \mathrm{d}\rho \;\Rightarrow\; \frac{\delta F}{\delta \rho} = U,
  \qquad
  F(\rho) = \int \rho \log \rho \;\Rightarrow\; \frac{\delta F}{\delta \rho} = \log \rho + 1,
  \nonumber
\end{equation}
where $U : \mathcal{X} \to \mathbb{R}$ is a potential and, in the second case, $\rho$ also denotes its density.

\paragraph{Wasserstein gradient flow.}
A curve $(\rho_t)_{t \ge 0}$ in $\mathcal{P}_2(\mathcal{X})$ is generated by a velocity field $v_t : \mathcal{X} \to \mathcal{X}$ when mass is conserved along it, that is, when the continuity equation
\begin{equation}
  \partial_t \rho_t + \nabla \cdot (\rho_t v_t) = 0
  \nonumber
\end{equation}
holds; each particle then moves by $\dot{x}_t = v_t(x_t)$.
The Benamou--Brenier formula~\citep{benamou2000computational} identifies $W_2$ with the minimal kinetic energy of such curves,
\begin{equation}
  W_2^2(\rho_0, \rho_1) = \min_{(\rho_t, v_t)} \int_0^1 \!\! \int_{\mathcal{X}} \|v_t(x)\|^2 \, \mathrm{d}\rho_t(x) \, \mathrm{d}t ,
  \nonumber
\end{equation}
so the speed of the curve in $W_2$ is $\|v_t\|_{L^2(\rho_t)}$.
Along the curve, the definition of the first variation and integration by parts give
\begin{equation}
  \frac{\mathrm{d}}{\mathrm{d}t} F(\rho_t)
  = \int \frac{\delta F}{\delta \rho}(\rho_t) \, \partial_t \rho_t
  = \int \Big\langle \nabla \frac{\delta F}{\delta \rho}(\rho_t), \, v_t \Big\rangle \, \mathrm{d}\rho_t .
  \nonumber
\end{equation}
Among velocities of a given norm $\|v_t\|_{L^2(\rho_t)}$, the one that decreases $F$ fastest is $v_t = -\nabla \frac{\delta F}{\delta \rho}(\rho_t)$.
The Wasserstein gradient flow of $F$~\citep{ambrosio2008gradient, santambrogio2015optimal} is the curve driven by this steepest-descent velocity:
\begin{equation}
  \partial_t \rho_t = \nabla \cdot \Big( \rho_t \, \nabla \frac{\delta F}{\delta \rho}(\rho_t) \Big).
  \label{eq:wgf-supp}
\end{equation}
For the potential energy $F(\rho) = \int U \, \mathrm{d}\rho$, every particle follows $\dot{x}_t = -\nabla U(x_t)$ independently of the others.
For $F(\rho) = \int U \, \mathrm{d}\rho + \int \rho \log \rho$, the identity $\nabla \cdot (\rho \nabla \log \rho) = \Delta \rho$ gives the Fokker--Planck equation $\partial_t \rho_t = \nabla \cdot (\rho_t \nabla U) + \Delta \rho_t$.

\paragraph{JKO scheme.}
The JKO scheme~\citep{jordan1998variational} discretizes the flow in time with step size $\tau > 0$:
\begin{equation}
  \rho_{k+1} = \arg\min_{\rho \in \mathcal{P}_2(\mathcal{X})} \; \frac{1}{2\tau} W_2^2(\rho_k, \rho) + F(\rho).
  \label{eq:jko-supp}
\end{equation}
It is the Wasserstein analogue of the proximal (implicit Euler) step $x_{k+1} = \arg\min_x \frac{1}{2\tau}\|x - x_k\|^2 + f(x)$ in $\mathbb{R}^n$.
Its optimality condition makes this precise: if $T_k$ is the optimal map from $\rho_{k+1}$ to $\rho_k$, then
\begin{equation}
  T_k(x) = x + \tau \, \nabla \frac{\delta F}{\delta \rho}(\rho_{k+1})(x) \qquad \rho_{k+1}\text{-a.e.},
  \nonumber
\end{equation}
so each point moves from $T_k(x)$ to $x$ by one step of $-\nabla \frac{\delta F}{\delta \rho}$ evaluated at the new measure.
For the Fokker--Planck case, the piecewise-constant interpolation of $(\rho_k)$ converges to the gradient flow as $\tau \to 0$~\citep{jordan1998variational}, and the convergence extends to general $\lambda$-convex functionals~\citep{ambrosio2008gradient}.

\paragraph{Map form.}
Restricting the new measure to a pushforward $\rho = T_\# \rho_k$ and using the coupling $(\mathrm{id}, T)_\# \rho_k$ gives $W_2^2(\rho_k, T_\# \rho_k) \le \int \|x - T(x)\|^2 \mathrm{d}\rho_k$, with equality when $T$ is optimal.
This yields the map form of the JKO step~\citep{mokrov2021large, fan2021variational, alvarezmelis2022optimizing},
\begin{equation}
  T_k = \arg\min_{T} \; \int \frac{\|x - T(x)\|^2}{2\tau} \, \mathrm{d}\rho_k(x) + F(T_\# \rho_k),
  \qquad \rho_{k+1} = (T_k)_\# \rho_k ,
  \nonumber
\end{equation}
where $T$ is parameterized by a neural network and both terms are estimated from samples of $\rho_k$.

For a potential energy the JKO step has a closed form: it decouples over points and is a c-transform.

\begin{lemma}[JKO step of a potential energy]
\label{lem:jko_linear}
Let $F(\rho) = \int U \, \mathrm{d}\rho$ and $c_\tau(x,y) = \|x-y\|^2 / (2\tau)$.
Assume that for $\rho_k$-a.e.\ $x$ the infimum
\begin{equation}
  (-U)^{c_\tau}(x) = \inf_{y \in \mathcal{X}} \big[ c_\tau(x, y) + U(y) \big]
  \nonumber
\end{equation}
is attained, and let $T : \mathcal{X} \to \mathcal{X}$ be a measurable selection $T(x) \in \arg\min_y [c_\tau(x,y) + U(y)]$.
Then $\rho_{k+1} = T_\# \rho_k$ solves the JKO step, $T$ solves its map form, and the optimal value is $\int (-U)^{c_\tau} \mathrm{d}\rho_k$.
\end{lemma}
\begin{proof}
For any $\rho$ and any $\gamma \in \Pi(\rho_k, \rho)$,
\begin{equation}
  \int c_\tau \, \mathrm{d}\gamma + \int U \, \mathrm{d}\rho
  = \int \big[ c_\tau(x,y) + U(y) \big] \mathrm{d}\gamma(x,y)
  \ge \int (-U)^{c_\tau}(x) \, \mathrm{d}\rho_k(x).
  \nonumber
\end{equation}
Taking the infimum over $\gamma$ gives $\frac{1}{2\tau} W_2^2(\rho_k, \rho) + F(\rho) \ge \int (-U)^{c_\tau} \mathrm{d}\rho_k$ for every $\rho$.
The coupling $(\mathrm{id}, T)_\# \rho_k$ attains equality, so $T_\# \rho_k$ attains the lower bound in both the measure form and the map form.
\end{proof}

\subsection{Reinforcement learning as a Wasserstein gradient flow}
\label{app:bg_rl_wgf}

Fix a state $s$ and view the action distribution $\pi(\cdot \mid s) \in \mathcal{P}_2(\mathcal{A})$ as the object being optimized.
By Lemma~\ref{lem:pdl}, keeping the state distribution of $\pi_{\mathrm{old}}$ fixed, the return is linear in each $\pi(\cdot \mid s)$:
\begin{equation}
  J(\pi) \approx J(\pi_{\mathrm{old}})
  + \frac{1}{1-\gamma} \, \mathbb{E}_{s \sim d^{\pi_{\mathrm{old}}}}
  \Big[ \int_{\mathcal{A}} A^{\pi_{\mathrm{old}}}(s, a) \, \mathrm{d}\pi(a \mid s) \Big],
  \nonumber
\end{equation}
which is the TRPO surrogate of Section~\ref{app:bg_trpo}.
At each state, maximizing this objective is minimizing the potential energy with $U = -A^{\pi_{\mathrm{old}}}(s, \cdot)$, so its Wasserstein gradient flow moves every action along the action-gradient of the advantage,
\begin{equation}
  \partial_t \pi_t(\cdot \mid s) = -\nabla_a \cdot \big( \pi_t(\cdot \mid s) \, \nabla_a A^{\pi_t}(s, \cdot) \big),
  \qquad
  \dot{a}_t = \nabla_a A^{\pi_t}(s, a_t) = \nabla_a Q^{\pi_t}(s, a_t).
  \nonumber
\end{equation}
This view underlies a line of policy-optimization methods.
\citet{richemond2017wasserstein} and \citet{zhang2018policy} write entropy-regularized policy optimization as a Wasserstein gradient flow (a Fokker--Planck equation in action space) and discretize it with JKO;
\citet{ziesche2023wasserstein} apply it to Gaussian-mixture policies;
\citet{terpin2022trust} and \citet{song2023provably} replace the KL trust region of TRPO by an optimal-transport discrepancy;
WPO~\citep{pfau2025wasserstein} projects the action-space flow onto the network parameters;
and WPPG~\citep{zhu2026wasserstein} takes the JKO step with an implicit (pushforward) policy $\pi_\theta(\cdot \mid s) = g_\theta(s, \cdot)_\# \nu$, updating the generator by
\begin{equation}
  \max_{\theta} \; \mathbb{E}_{s,\, z \sim \nu} \Big[ Q(s, g_\theta(s, z)) - \frac{1}{2\tau} \big\| g_\theta(s, z) - g_{\mathrm{old}}(s, z) \big\|^2 \Big].
  \nonumber
\end{equation}
For flow policies, SWFP~\citep{sun2025iterative} treats each block of the flow as one JKO step with the soft Q-function as energy.

The JKO step with the advantage as potential is the penalty form of a $W_2$ trust region:
\begin{equation}
  \pi_{k+1}(\cdot \mid s) = \arg\min_{\pi} \; \frac{1}{2\tau} W_2^2\big( \pi_k(\cdot \mid s), \pi \big) - \int_{\mathcal{A}} A^{\pi_k}(s, a) \, \mathrm{d}\pi(a),
  \nonumber
\end{equation}
and by Lemma~\ref{lem:jko_linear} it is the pushforward of $\pi_k(\cdot \mid s)$ by the c-transform map
\begin{equation}
  T_s(a) = \arg\max_{a' \in \mathcal{A}} \Big[ A^{\pi_k}(s, a') - \frac{\|a - a'\|^2}{2\tau} \Big].
  \nonumber
\end{equation}
All methods above evaluate $Q$, $A$, or $\nabla_a Q$ at arbitrary actions and therefore require a learned critic.

\section{Method: soft c-transform GRPO}
\label{app:method}

The method keeps the GRPO objective of Section~\ref{app:bg_flowgrpo} and adds one regression term.
The term is derived in four steps:
the JKO step of Section~\ref{app:bg_rl_wgf} (Section~\ref{app:m_jko}),
its restriction to the samples of one GRPO group, which removes the critic (Section~\ref{app:m_group}),
its entropic smoothing (Section~\ref{app:m_soft}),
and its projection onto the flow-matching policy (Section~\ref{app:m_map}).
Section~\ref{app:m_impl} lists the choices made for VLA policies, and Section~\ref{app:m_obj} states the full objective.

\paragraph{Remark (correspondence with the main text).}
\Cref{lem:main_group_jko} of the main text is split here into \cref{lem:m_group_jko}, which gives its part (i), and \cref{cor:m_move}, which gives its part (ii).
The limit $y_i \to T(x_i)$ as $\kappa \to 0$ stated after \eqref{eq:ct-target} is \cref{lem:m_soft}(ii), and the exact-projection property stated after \eqref{eq:ct-loss} is \cref{lem:m_proj}.
We refer the related proofs to these statements.

\subsection{Starting point: the JKO step of the advantage}
\label{app:m_jko}

Fix a state $s \in \mathcal{S}$ and let $\pi_{\mathrm{old}}(\cdot \mid s) \in \mathcal{P}_2(\mathcal{A})$ be the current action distribution, with $\mathcal{A} = \mathbb{R}^{H \times d}$ the space of action chunks.
For a step size $\tau > 0$, write
\begin{equation}
  c_\tau : \mathcal{A} \times \mathcal{A} \to \mathbb{R}_{\ge 0}, \qquad
  c_\tau(a, a') = \frac{\|a - a'\|^2}{2\tau},
  \nonumber
\end{equation}
where $\|\cdot\|$ is the Frobenius norm on $\mathbb{R}^{H \times d}$.
The JKO step of Section~\ref{app:bg_rl_wgf} with potential $U = -A^{\pi_{\mathrm{old}}}(s, \cdot)$ is
\begin{equation}
\label{eq:m_jko}
  \pi_{\mathrm{new}}(\cdot \mid s)
  = \arg\min_{\rho \in \mathcal{P}_2(\mathcal{A})}
  \; \inf_{\gamma \in \Pi(\pi_{\mathrm{old}}(\cdot \mid s), \rho)} \int c_\tau \, \mathrm{d}\gamma
  \;-\; \int_{\mathcal{A}} A^{\pi_{\mathrm{old}}}(s, a) \, \mathrm{d}\rho(a),
\end{equation}
and by Lemma~\ref{lem:jko_linear} its solution moves each action $a$ to
\begin{equation}
  T_s(a) = \arg\max_{a' \in \mathcal{A}} \big[ A^{\pi_{\mathrm{old}}}(s, a') - c_\tau(a, a') \big], \pi_{new}=(T_s)_\# \pi_{old}(\cdot|s)\nonumber 
\end{equation}
Evaluating $T_s$ needs $A^{\pi_{\mathrm{old}}}(s, a')$ at every $a' \in \mathcal{A}$, i.e.\ a learned critic.
GRPO provides the advantage only at the actions it has sampled; the next step uses exactly these values.

\subsection{Group-supported JKO step}
\label{app:m_group}

\paragraph{The group.}
GRPO samples $G$ actions from $\pi_{\mathrm{old}}(\cdot \mid s)$ at the same state $s$.
For the sampler \eqref{eq:sampler}, each sample is the image of a noise sequence:
\begin{equation}
  \boldsymbol{\epsilon}_i \in \mathcal{A}^{K+1}, \quad \boldsymbol{\epsilon}_i \sim \mu_K, \qquad
  x_i = f_{\mathrm{old}}(\boldsymbol{\epsilon}_i; s) \in \mathcal{A}, \qquad
  A_i \in \mathbb{R}, \qquad i = 1, \dots, G,
  \nonumber
\end{equation}
where $f_{\mathrm{old}} = f_{\theta_{\mathrm{old}}}$ and $A_i$ is the group-relative advantage of Section~\ref{app:bg_grpo}.
The group defines the finite candidate set $X_G = \{x_1, \dots, x_G\} \subset \mathcal{A}$, the empirical measure
\begin{equation}
  \hat{\rho} = \frac{1}{G} \sum_{i=1}^{G} \delta_{x_i} \in \mathcal{P}(X_G),
  \nonumber
\end{equation}
and the advantage $\hat{A} : X_G \to \mathbb{R}$, $\hat{A}(x_j) = A_j$, which is known on all of $X_G$.

\paragraph{Restricted JKO step.}
Replacing $\pi_{\mathrm{old}}(\cdot \mid s)$ by $\hat{\rho}$ and the search space $\mathcal{P}_2(\mathcal{A})$ by $\mathcal{P}(X_G)$ in \eqref{eq:m_jko} gives
\begin{equation}
\label{eq:m_group_jko}
  \min_{\rho \in \mathcal{P}(X_G)} \; \mathcal{J}(\rho), \qquad
  \mathcal{J}(\rho) = \inf_{\gamma \in \Pi(\hat{\rho}, \rho)} \int c_\tau \, \mathrm{d}\gamma \;-\; \int \hat{A} \, \mathrm{d}\rho .
\end{equation}
Every quantity in \eqref{eq:m_group_jko} is available from the rollouts.

\begin{lemma}[Group-supported JKO step]
\label{lem:m_group_jko}
Assume the sampled action locations are distinct. Define $T : X_G \to X_G$, breaking ties in favor of $x_i$ when possible, by
\begin{equation}
\label{eq:m_hard_ct}
  T(x_i) \in \arg\max_{j \in \{1, \dots, G\}} \big[ A_j - c_\tau(x_i, x_j) \big], \qquad i = 1, \dots, G.
\end{equation}
Then $T_\# \hat{\rho}$ solves \eqref{eq:m_group_jko}, and the optimal value is
$\frac{1}{G} \sum_{i} \min_j \big[ c_\tau(x_i, x_j) - A_j \big]$, the c-transform of $\hat{A}$ averaged over $\hat{\rho}$.
\end{lemma}
\begin{proof}
For any $\rho \in \mathcal{P}(X_G)$ and $\gamma \in \Pi(\hat{\rho}, \rho)$, $\gamma$ is supported on $X_G \times X_G$, and
\begin{equation}
  \int c_\tau \, \mathrm{d}\gamma - \int \hat{A} \, \mathrm{d}\rho
  = \int \big[ c_\tau(x, y) - \hat{A}(y) \big] \mathrm{d}\gamma(x, y)
  \ge \int \min_{y \in X_G} \big[ c_\tau(x, y) - \hat{A}(y) \big] \mathrm{d}\hat{\rho}(x).
  \nonumber
\end{equation}
The coupling $(\mathrm{id}, T)_\# \hat{\rho} \in \Pi(\hat{\rho}, T_\# \hat{\rho})$ attains equality. This is Lemma~\ref{lem:jko_linear} with $\mathcal{X}$ replaced by the finite set $X_G$, on which the minimum is always attained.
\end{proof}

Lemma~\ref{lem:m_group_jko} gives a simple condition for a sample to move.

\begin{corollary}[When a sample moves]
\label{cor:m_move}
Since $c_\tau(x_i, x_i) = 0$, the candidate $j = i$ always has value $A_i$.
Hence $T(x_i) \neq x_i$ only if there is a $j$ with
\begin{equation}
  A_j - A_i > \frac{\|x_i - x_j\|^2}{2\tau}.
  \nonumber
\end{equation}
A sample moves to another member of its group only when the advantage gain exceeds the transport cost; $\tau$ sets the exchange rate between the two and is the proximal step size.
If all $A_j$ in a group are equal, $T = \mathrm{id}$ and the group carries no signal, as in GRPO.
\end{corollary}

\subsection{Soft c-transform targets}
\label{app:m_soft}

The argmax in \eqref{eq:m_hard_ct} is discontinuous in $(A_j, x_j)$ and selects a single member.
Applying the soft c-transform of Section~\ref{app:bg_ot_basic} with $\psi = \hat{A}$, $b_j = 1/G$ and temperature $\kappa > 0$ replaces it by Gibbs weights over the group,
\begin{equation}
\label{eq:m_weights}
  p_{ij} = \frac{\exp\big( (A_j - c_\tau(x_i, x_j)) / \kappa \big)}{\sum_{l=1}^{G} \exp\big( (A_l - c_\tau(x_i, x_l)) / \kappa \big)},
  \qquad i, j = 1, \dots, G,
\end{equation}
and the target of sample $i$ is the mean of the group under these weights,
\begin{equation}
\label{eq:m_target}
  y_i = \sum_{j=1}^{G} p_{ij} \, x_j \in \mathcal{A}.
\end{equation}
The problem \eqref{eq:m_group_jko} constrains only the first marginal of $\gamma$, so the weights are a single softmax per row and require no Sinkhorn iterations.

\begin{lemma}[Properties of the soft target]
\label{lem:m_soft}
\begin{enumerate}
  \item[(i)] $y_i$ lies in the convex hull of $X_G$.
  \item[(ii)] If the maximizer in \eqref{eq:m_hard_ct} is unique, then $y_i \to T(x_i)$ as $\kappa \to 0$.
  \item[(iii)] $p_{ii}$ is the weight that sample $i$ keeps on itself. If $p_{ii} \ge 1/2$, then $j = i$ maximizes $A_j - c_\tau(x_i, x_j)$, so $T(x_i) = x_i$ can be chosen in \eqref{eq:m_hard_ct}. If the maximizer in \eqref{eq:m_hard_ct} is unique, then $\mathbf{1}[p_{ii} < 1/2] \to \mathbf{1}[T(x_i) \neq x_i]$ as $\kappa \to 0$.
\end{enumerate}
\end{lemma}
\begin{proof}
(i) The weights are nonnegative and sum to one.

(ii) As $\kappa \to 0$ the softmax concentrates on the unique maximizer $j^\star$ of $A_j - c_\tau(x_i, x_j)$, so $p_{ij} \to \mathbb{1}[j = j^\star]$ and $y_i \to x_{j^\star} = T(x_i)$.

(iii) Write $z_{ij} = A_j - c_\tau(x_i, x_j)$, so that $p_{ij} = e^{z_{ij}/\kappa} / \sum_l e^{z_{il}/\kappa}$ and $z_{ii} = A_i$. Since $\sum_j p_{ij} = 1$, $p_{ii} \ge 1/2$ gives $p_{ii} \ge 1 - p_{ii} = \sum_{j \neq i} p_{ij} \ge p_{ij}$ for every $j$. The weights of row $i$ share the denominator, so $p_{ii} \ge p_{ij}$ is $e^{z_{ii}/\kappa} \ge e^{z_{ij}/\kappa}$, i.e.\ $A_i \ge A_j - c_\tau(x_i, x_j)$ for every $j$; hence $j = i$ attains the maximum in \eqref{eq:m_hard_ct}. For the limit, by (ii), $p_{ii} \to 1$ if $i$ is the unique maximizer, i.e.\ $T(x_i) = x_i$, and $p_{ii} \to 0$ otherwise, i.e.\ $T(x_i) \neq x_i$.
\end{proof}

\paragraph{Connection to the c-transform target.}
The Fisher--Rao and transport mechanisms of \cref{app:bg_flowgrpo_fr} appear as limits of the soft weights \eqref{eq:m_weights} and targets \eqref{eq:m_target} in $\tau$, at fixed $\kappa$.
\begin{enumerate}
\item As $\tau \to \infty$, $c_\tau(x_i, x_j) \to 0$ for all $i, j$, so $p_{ij} \to e^{A_j/\kappa} / \sum_{l} e^{A_l/\kappa}$ for every $i$. These are the weights of the Fisher--Rao step \eqref{eq:fr_closed} on the empirical measure $\frac{1}{G}\sum_i \delta_{x_i}$ with $\alpha = 1/\kappa$, and every target $y_i$ tends to the mean of the reweighted group measure.
\item As $\tau \to 0$, $c_\tau(x_i, x_j) \to \infty$ for $x_j \neq x_i$, so $p_{ij} \to \mathbf{1}\{j = i\}$ for distinct samples and $y_i \to x_i$; no sample moves.
\end{enumerate}
For finite $\tau$, $p_{ij}$ favors group members that are both close to $x_i$ and of high advantage, so each sample moves toward nearby better actions. This transport behavior is absent from a Fisher--Rao step, and $\tau$ sets the balance between the two mechanisms.

\paragraph{Relation to the advantage gradient.}
For a differentiable advantage on $\mathcal{A}$, the first-order condition of $\max_{a'} [A(s, a') - c_\tau(a, a')]$ is
\begin{equation}
  T_s(a) = a + \tau \, \nabla_a A^{\pi_{\mathrm{old}}}\big(s, T_s(a)\big),
  \nonumber
\end{equation}
the implicit (proximal) step of the flow $\dot{a} = \nabla_a A$ in Section~\ref{app:bg_rl_wgf}.
The displacement $y_i-x_i$ provides a group-based surrogate direction. At fixed group size, shrinking $\tau$ to zero yields zero displacement for distinct sampled actions.

\subsection{Projection onto the flow-matching policy}
\label{app:m_map}

The targets $y_i$ are points in $\mathcal{A}$; the policy is the pushforward $\pi_\theta(\cdot \mid s) = f_\theta(\cdot\,; s)_\# \mu_K$ of the noise law by the random mapping \eqref{eq:sampler}.
Following the map form of the JKO step (Section~\ref{app:bg_wgf}), the new policy is coupled to the old one through the shared noise, $\boldsymbol{\epsilon} \mapsto (f_{\mathrm{old}}(\boldsymbol{\epsilon}; s), f_\theta(\boldsymbol{\epsilon}; s))$, and $f_\theta(\boldsymbol{\epsilon}_i; s)$ is asked to land on the target of the sample that $\boldsymbol{\epsilon}_i$ produced:
\begin{equation}
\label{eq:m_loss_group}
  \mathcal{L}_{\mathrm{ct}}(\theta; s)
  = \frac{1}{G} \sum_{i=1}^{G} m_i \, \big\| f_\theta(\boldsymbol{\epsilon}_i; s) - \mathrm{sg}(y_i) \big\|^2,
\end{equation}
where $\mathrm{sg}$ is the stop-gradient and $m_i \in \{0, 1\}$ is a mask defined in Section~\ref{app:m_impl}.

\begin{lemma}[Exact projection]
\label{lem:m_proj}
If $m_i = 1$ for all $i$ and $\mathcal{L}_{\mathrm{ct}}(\theta; s) = 0$, then the empirical pushforward of the group noises by the new policy equals the soft c-transform image of the group,
\begin{equation}
  \frac{1}{G} \sum_{i=1}^{G} \delta_{f_\theta(\boldsymbol{\epsilon}_i; s)} = \frac{1}{G} \sum_{i=1}^{G} \delta_{y_i},
  \nonumber
\end{equation}
which as $\kappa \to 0$ is the solution $T_\# \hat{\rho}$ of the group-supported JKO step (Lemmas~\ref{lem:m_group_jko} and~\ref{lem:m_soft}(ii)).
\end{lemma}
\begin{proof}
$\mathcal{L}_{\mathrm{ct}} = 0$ gives $f_\theta(\boldsymbol{\epsilon}_i; s) = y_i$ for every $i$.
\end{proof}

\paragraph{Gradient.}
The gradient of \eqref{eq:m_loss_group} is
\begin{equation}
  \nabla_\theta \mathcal{L}_{\mathrm{ct}}
  = \frac{2}{G} \sum_{i=1}^{G} m_i \, \big[ \partial_\theta f_\theta(\boldsymbol{\epsilon}_i; s) \big]^{\top} \big( f_\theta(\boldsymbol{\epsilon}_i; s) - y_i \big),
  \nonumber
\end{equation}
where $\partial_\theta f_\theta(\boldsymbol{\epsilon}_i; s)$ is the Jacobian of the sampler with respect to the parameters.
It uses the samples and the Jacobian of the sampler at the stored noise only, so it needs neither the density $\pi_\theta(a \mid s)$ nor a Gaussian transition.
At $\theta = \theta_{\mathrm{old}}$, $f_\theta(\boldsymbol{\epsilon}_i; s) = x_i$ and the negative gradient is $2\tau$ times the masked empirical pathwise update using $(y_i-x_i)/\tau$ as its action-space direction.

\subsection{Choices for VLA policies}
\label{app:m_impl}

\paragraph{Sampler.}
Rollouts use the hybrid sampler \eqref{eq:sampler} of $\pi_{\texttt{RL}}$~\cite{pirl2026}: $\sigma_{u_k} > 0$ at one randomly chosen denoising step per chunk; the other steps are Euler steps of the ODE (\cref{app:bg_flowgrpo}). The main transport loss uses deterministic ODE replay from the stored initial noise. A hybrid ODE--SDE replay using the stored stochastic-step noise that reproduces the executed chunk is compared in \cref{tab:supp-replay}.
The gradient of \eqref{eq:m_loss_group} is back-propagated through all $K$ steps, with activation checkpointing per step.
At evaluation, the policy uses the deterministic ODE sampler, as in Flow-GRPO.

\paragraph{Which chunk.}
The $G$ chunks of a group must share the observation $s$, which an episode of $\pi_{\mathrm{old}}$ provides only at its initial observation $s_0$.
We therefore use the random-time branching of \cref{sec:multi}: at each branch index $h_m$ of an episode,
 with $h_1, \dots, h_M$ drawn uniformly and without replacement from $\{0, \dots, H-1\}$, the simulator state is restored and $G$ continuations are rolled out from $s_{h_m}$ with $\pi_{\mathrm{old}}$.
The term \eqref{eq:m_loss_group} is applied with $s = s_{h_m}$ to the branch chunks $x_i = f_{\mathrm{old}}(\boldsymbol{\epsilon}_i; s_{h_m})$, where $\boldsymbol{\epsilon}_i$ is the noise sequence stored for the branch chunk of continuation $i$, and $A_i$ is computed by \eqref{eq:adv} from the returns $R_i$ of the continuations.

\paragraph{Coordinates.}
Actions are compared in the normalized space in which $v_\theta$ is trained, restricted to the $L$ executed steps and the $d$ environment dimensions, and $\|\cdot\|^2$ sums over all $L \times d$ entries.
The targets $x_i$, $p_{ij}$, $y_i$ are computed once per rollout batch with $\theta_{\mathrm{old}}$ and no gradient.

\paragraph{Mask.}
Groups with $\max_j A_j = \min_j A_j$ are skipped (Corollary~\ref{cor:m_move}), and within the remaining groups only samples that move contribute (Lemma~\ref{lem:m_soft}(iii)):
\begin{equation}
  m_i = \mathbb{1}\big[ \max_j A_j > \min_j A_j \big] \cdot \mathbb{1}\big[ p_{ii} < \tfrac12 \big].
  \nonumber
\end{equation}

\subsection{Full objective}
\label{app:m_obj}

For a branch group with shared observation $s = s_{h_m}$, the policy is trained on the objective \eqref{eq:total} of the main text,
\begin{equation}
\label{eq:m_full}
\mathcal{L}(\theta) = \lambda_1 \mathcal{L}_{\mathrm{ct}}(\theta; s) -\lambda_2 \mathcal{J}_{\mathrm{FG}}(\theta),
\end{equation}
averaged over all $NM$ branch groups of the rollout batch.
Here $\mathcal{J}_{\mathrm{FG}}$ is the clipped Flow-GRPO objective \eqref{eq:flowgrpo} with $\beta = 0$, evaluated on chunks $h_m, \dots, H-1$ of each continuation, where, as for the baseline, the sum over denoising steps in \eqref{eq:flowgrpo} has a single term per chunk, at its stochastic step, $\mathcal{L}_{\mathrm{ct}}$ is \eqref{eq:m_loss_group} on the branch chunks, and $\lambda_1 > 0$ weights the soft c-transform term.
The method has three hyper-parameters: $\tau$, which sets how far a sample may move per update (Corollary~\ref{cor:m_move}); $\kappa$, which interpolates between the hard c-transform ($\kappa \to 0$) and a uniform average over the group; and $\lambda_1$.
Setting $\lambda_1=0$ and $M=0$ recovers the Flow-GRPO baseline without the KL penalty.
In the reading of \cref{app:bg_flowgrpo_fr}, the likelihood-ratio term of \eqref{eq:m_full} supplies the Fisher--Rao component, and the c-transform term supplies the transport component.

\section{Related Work: RL in Flow Matching, VLA and OT}
\label{app:related}

\subsection{RL for diffusion, flow-matching and VLA policies}

\paragraph{DDPO.}
DDPO~\cite{black2024ddpo} casts the $K$-step denoising process of a diffusion model as an MDP: the state is $(x^{(k)}, u_k, s)$, where $x^{(k)}$ is the noisy sample at denoising step $k$, $u_k$ its time and $s$ the prompt; the action is the next sample $x^{(k+1)}$; the policy is the Gaussian reverse transition $\pi_\theta(x^{(k+1)} \mid x^{(k)}, s)$; and the reward $R(x^{(K)}, s)$ is given only to the final sample $x^{(K)}$.
It maximizes the expected reward $J(\theta) = \mathbb{E}_{x^{(0:K)} \sim \pi_\theta}[R(x^{(K)}, s)]$, where $x^{(0:K)}$ is a denoising chain generated by $\pi_\theta$, with two estimators of $\nabla_\theta J$:
\begin{align}
\nabla_\theta J_{\mathrm{SF}} &= \mathbb{E}_{\pi_\theta}\Big[\sum_{k=0}^{K-1} \nabla_\theta \log \pi_\theta\big(x^{(k+1)} \mid x^{(k)}, s\big)\, R(x^{(K)}, s)\Big], \nonumber \\
\nabla_\theta J_{\mathrm{IS}} &= \mathbb{E}_{\pi_{\mathrm{old}}}\Big[\sum_{k=0}^{K-1} \frac{\pi_\theta(x^{(k+1)} \mid x^{(k)}, s)}{\pi_{\mathrm{old}}(x^{(k+1)} \mid x^{(k)}, s)}\, \nabla_\theta \log \pi_\theta\big(x^{(k+1)} \mid x^{(k)}, s\big)\, R(x^{(K)}, s)\Big]. \nonumber
\end{align}
The first is the REINFORCE (score-function) estimator on chains of the current policy; the second reuses chains of the rollout policy $\pi_{\mathrm{old}}$ through the per-step likelihood ratio, which is clipped as in PPO.
Both need the closed-form density of each Gaussian transition.
DDPO fine-tunes text-to-image models on rewards such as JPEG compressibility, aesthetic score and prompt--image alignment.

\paragraph{DPOK.}
DPOK~\cite{fan2023dpok} adds to the objective of DDPO a KL penalty toward the pretrained model $\pi_{\mathrm{pre}}$.
The KL divergence between the distributions $\pi_\theta(x^{(K)} \mid s)$ and $\pi_{\mathrm{pre}}(x^{(K)} \mid s)$ of final samples has no closed form, and DPOK replaces it by its upper bound, the sum of the per-step KL divergences between Gaussian transitions, which follows from the chain rule of the KL divergence on the whole chain:
\begin{align}
\mathrm{KL}\big(\pi_\theta(x^{(K)} \mid s) \,\|\, \pi_{\mathrm{pre}}(x^{(K)} \mid s)\big) &\le \sum_{k=0}^{K-1} \mathbb{E}_{\pi_\theta}\Big[\mathrm{KL}\big(\pi_\theta(\cdot \mid x^{(k)}, s) \,\|\, \pi_{\mathrm{pre}}(\cdot \mid x^{(k)}, s)\big)\Big], \nonumber \\
\mathcal{L}_{\mathrm{DPOK}}(\theta) &= \mathbb{E}_{s}\Big[-\mathbb{E}_{\pi_\theta}\big[R(x^{(K)}, s)\big] \nonumber\\
&\qquad + \beta \sum_{k=0}^{K-1} \mathbb{E}_{\pi_\theta}\Big[\mathrm{KL}\big(\pi_\theta(\cdot \mid x^{(k)}, s) \,\|\, \pi_{\mathrm{pre}}(\cdot \mid x^{(k)}, s)\big)\Big]\Big], \nonumber
\end{align}
where $\beta > 0$ weights the penalty.
Each per-step KL is between two Gaussians and is computed in closed form; the reward term is optimized by the policy gradient of DDPO.
The reference penalty of Flow-GRPO (\cref{app:bg_flowgrpo}) has the same form.
In the Fisher--Rao reading (\cref{app:bg_fisher_rao}), this penalty changes the functional from $-\mathbb{E}_P[A]$ to $-\mathbb{E}_P[A] + \beta\, \mathrm{KL}(P \,\|\, \pi_{\mathrm{ref}})$ and keeps the geometry; our method uses $\beta = 0$ and adds a transport ($W_2$) component to the update.

\paragraph{Flow-GRPO.}
Flow-GRPO~\cite{liu2026flow} applies GRPO to flow-matching models.
A flow-matching model samples by the deterministic ODE $\mathrm{d}x = v_\theta(x, u; s)\, \mathrm{d}u$, which has no transition density.
Flow-GRPO replaces it by an SDE with the same marginals $p_u$, whose drift is $v_\theta + \frac{\sigma_u^2}{2} \nabla \log p_u$ with noise level $\sigma_u$ and the score $\nabla \log p_u$ written through $v_\theta$ (Lemma~\ref{lem:flow_sde}); each Euler--Maruyama step is then a Gaussian transition $\pi_\theta(x^{(k+1)} \mid x^{(k)}, s)$ with a closed-form density.
For a group of $G$ chains $\{x_i^{(0:K)}\}_{i=1}^{G}$ generated from the same $s$ by $\pi_{\mathrm{old}}$, it maximizes
{\begin{align}
J_{\mathrm{Flow\text{-}GRPO}}(\theta) &= \mathbb{E}\bigg[\frac{1}{G}\sum_{i = 1}^{G} \frac{1}{K}\sum_{k = 0}^{K-1} \min\big(w_{i,k}(\theta) A_i,\; \mathrm{clip}(w_{i,k}(\theta), 1-\epsilon, 1+\epsilon)\, A_i\big) \nonumber\\
&\qquad - \beta\, \mathrm{KL}\big(\pi_\theta(\cdot \mid x_i^{(k)}, s) \,\|\, \pi_{\mathrm{ref}}(\cdot \mid x_i^{(k)}, s)\big)\bigg], \nonumber
\end{align}}
where $w_{i,k}(\theta) = \pi_\theta(x_i^{(k+1)} \mid x_i^{(k)}, s) / \pi_{\mathrm{old}}(x_i^{(k+1)} \mid x_i^{(k)}, s)$ is the per-step likelihood ratio, $A_i$ is the group-relative advantage of the final sample $x_i^{(K)}$ (the reward minus the group mean, divided by the group standard deviation), $\epsilon$ is the clipping range, and $\beta$ weights the per-step KL toward the reference model $\pi_{\mathrm{ref}}$ (\cref{app:bg_flowgrpo}).
Every denoising step of chain $i$ receives the same advantage $A_i$.
Flow-GRPO was developed for text-to-image generation, where one prompt yields one image and the reward is assigned to the final image.
Our objective \eqref{eq:m_full} keeps this likelihood-ratio term and adds the c-transform term.

\paragraph{Remark (KL versus OT gradient flow).}
An idealized version of the Flow-GRPO update is a KL-proximal step around $\pi_{\mathrm{old}}$ of the Fisher--Rao gradient flow of $-\mathbb{E}_P[A]$ (\cref{app:bg_flowgrpo_fr}): each update reweights actions that the old policy already produces.
Our c-transform step is a $W_2$-proximal (JKO) step of the same functional restricted to the group, and it moves each sample toward nearby group members of higher advantage.
The comparison of the two geometries is given in the remark of \cref{app:bg_flowgrpo_fr}, and the limits $\tau \to \infty$ and $\tau \to 0$, in which our targets reduce to Fisher--Rao reweighting and to no update, are given in \cref{app:m_soft}. JKO-GRPO keeps the former and adds the latter.

\paragraph{DanceGRPO.}
DanceGRPO~\cite{xue2025dancegrpo}, concurrent with Flow-GRPO, applies GRPO to both diffusion models and rectified flows.
Its key observation is that the deterministic samplers of the two model families are one update rule:
\begin{align}
\tilde z_{t'} = \tilde z_t + \hat f_\theta(z_t, t, s)\, \big(\eta_{t'} - \eta_t\big), \nonumber
\end{align}
where $z_t$ is the noisy sample at noise level $t$, $t' < t$ is the next, lower noise level, and $\hat f_\theta$ is the network output.
For a diffusion model with $z_t = \alpha_t x + \sigma_t \epsilon$, $\hat f_\theta$ is the noise prediction, $\tilde z_t = z_t / \alpha_t$ and $\eta_t = \sigma_t / \alpha_t$, which is the DDIM step; for a rectified flow with $z_t = (1-t) x + t \epsilon$, $\hat f_\theta$ is the velocity prediction, $\tilde z_t = z_t$ and $\eta_t = t$, which is the Euler step of the ODE.
DanceGRPO adds the noise of the corresponding reverse SDE to this update, so each step is a Gaussian transition, and optimizes on these transitions the clipped likelihood-ratio objective of Flow-GRPO above.
The samples of a group share the prompt and the initial noise, and only a subset of the denoising steps enters the objective.
It is evaluated on text-to-image, text-to-video and image-to-video generation.
Its update is the same idealized Fisher--Rao step as Flow-GRPO.

\paragraph{$\pi_{\mathrm{RL}}$.}
$\pi_{\mathrm{RL}}$~\cite{pirl2026} fine-tunes the flow-based VLAs $\pi_0$ and $\pi_{0.5}$ with online RL in parallel simulation.
Its Flow-SDE variant, which we build on, defines a two-layer MDP.
The outer MDP is the robot task: the state is the observation $o_t$ (images, language instruction and proprioception), the action is an action chunk $x_t \in \mathcal{A}$, and the environment returns the next observation $o_{t+1}$ and the chunk reward $R_t = \sum_{j=0}^{H-1} r_{t,j}$, the sum of the per-step rewards $r_{t,j}$ over the $H$ executed actions.
The inner MDP is the denoising process within one environment step: its state is $(o_t, x_t^{(k)})$, its action is the next denoised chunk $x_t^{(k+1)}$, drawn from the Gaussian transition of the Flow-GRPO SDE, and its reward is zero except at the last denoising step, where the chunk $x_t^{(K)}$ is executed and $R_t$ is received; the denoising state is then reset to fresh noise for $o_{t+1}$.
To shorten the horizon, $\pi_{\mathrm{RL}}$ makes one denoising step per environment step stochastic and runs the others as ODE steps, so the log-probability of a chunk is the log-density of that single Gaussian transition (the hybrid sampler of \cref{app:bg_flowgrpo}).
It estimates advantages by GAE with a learned critic and optimizes the PPO clipped objective, which is the objective of Flow-GRPO with $A_i$ replaced by the GAE advantage of each environment step.
Its Flow-Noise variant instead learns the injected noise level with a separate network and treats the whole denoising chain as one step of the outer MDP, with the joint log-likelihood of the chain.
Our robot experiments use the Flow-SDE setting of $\pi_{\mathrm{RL}}$ as base and main baseline, and add the c-transform term to its policy-gradient update.

\paragraph{DGPO.}
DGPO~\cite{luo2026dgpo} fine-tunes diffusion models from group rewards without a stochastic policy.
For each prompt $s$ it samples a group of $G$ final samples $x_1, \dots, x_G$ with the deterministic ODE sampler and computes the group-relative advantages $A_i$ as in Flow-GRPO.
The external reward enters only through $A_i$, which are fixed weights without gradient.
Following Diffusion-DPO~\cite{wallace2024diffusiondpo}, the reward of a final sample is defined implicitly by the model, through the denoising chains that end at that sample.
Let $p_\theta(x^{(0:K)} \mid s) = \mathcal{N}(x^{(0)}; 0, I) \prod_{k=0}^{K-1} \pi_\theta(x^{(k+1)} \mid x^{(k)}, s)$ be the density of a chain of the stochastic (Gaussian) reverse sampler of the model, $p_{\mathrm{ref}}$ that of the reference model, and $q(x^{(0:K-1)} \mid x^{(K)})$ the forward noising process, which maps a final sample $x^{(K)}$ back to noise.
The implicit reward of a final sample $x$ is
\begin{align}
r_\theta(s, x) = \beta\, \mathbb{E}_{q(x^{(0:K-1)} \mid x^{(K)} = x)}\Big[\log \frac{p_\theta(x^{(0:K)} \mid s)}{p_{\mathrm{ref}}(x^{(0:K)} \mid s)}\Big] + \beta \log Z(s), \nonumber
\end{align}
where the expectation is over chains ending at $x$, with $q$ used in place of the intractable posterior $p_\theta(x^{(0:K-1)} \mid x^{(K)}, s)$, $Z(s)$ is the intractable normalizer of the KL-regularized optimum $p_{\mathrm{ref}}\, e^{r/\beta}/Z$, and $\beta > 0$ is the KL weight.
DGPO splits the group into $\mathcal{G}^+ = \{x_i : A_i > 0\}$ and $\mathcal{G}^- = \{x_i : A_i \le 0\}$, gives each subgroup the reward $R_\theta(\mathcal{G}^\pm \mid s) = \sum_{x_i \in \mathcal{G}^\pm} |A_i|\, r_\theta(s, x_i)$, and maximizes the Bradley--Terry likelihood $\sigma\big(R_\theta(\mathcal{G}^+ \mid s) - R_\theta(\mathcal{G}^- \mid s)\big)$ that $\mathcal{G}^+$ is preferred to $\mathcal{G}^-$, with $\sigma$ the sigmoid.
Since $-|A_i| = A_i$ on $\mathcal{G}^-$ and $\sum_i A_i = 0$, the difference is $\sum_i A_i\, r_\theta(s, x_i)$, in which $\log Z(s)$ cancels:
\begin{align}
\mathcal{L}_{\mathrm{BT}}(\theta) = -\mathbb{E}_{s}\Big[\log \sigma\Big(\beta \sum_{i=1}^{G} A_i\, \mathbb{E}_{q(x_i^{(0:K-1)} \mid x_i)}\Big[\log \frac{p_\theta(x_i^{(0:K)} \mid s)}{p_{\mathrm{ref}}(x_i^{(0:K)} \mid s)}\Big]\Big)\Big]. \nonumber
\end{align}
The chain log-ratio is a sum over steps; under $q$, the term of each step equals the difference of the KL divergences from the forward posterior to the Gaussian transitions of $p_{\mathrm{ref}}$ and $p_\theta$, which is proportional to a difference of denoising losses.
Sampling one step and moving the expectation outside $-\log\sigma$ by Jensen's inequality gives the upper bound that DGPO minimizes,
\begin{align}
\mathcal{L}_{\mathrm{DGPO}}(\theta) &= -\mathbb{E}_{s, u, \epsilon}\Big[\log \sigma\Big(-\beta \lambda_u \sum_{i=1}^{G} A_i \big[\ell_\theta(x_i, x_{i,u}) - \ell_{\mathrm{ref}}(x_i, x_{i,u})\big]\Big)\Big], \nonumber \\
\ell_\theta(x, x_u) &= \big\|f_\theta(x_u, u, s) - x\big\|^2, \nonumber
\end{align}
where $u$ is a random noise level, $x_{i,u} = \alpha_u x_i + \sigma_u \epsilon$ is the noised sample with noise $\epsilon \sim \mathcal{N}(0, I)$ shared within the group, $f_\theta$ is the data prediction of the model and $f_{\mathrm{ref}}$ that of the reference model in $\ell_{\mathrm{ref}}$, and $\lambda_u > 0$ is a time weighting with the number of steps absorbed into $\beta$.
Minimizing $\mathcal{L}_{\mathrm{DGPO}}$ lowers the denoising loss of samples with $A_i > 0$ and raises it for samples with $A_i < 0$, in proportion to $|A_i|$.
The ODE sampler is used only to generate $x_i$; the loss treats them as data points and needs neither the rollout trajectory nor a transition density of the sampler.
Its target is the KL-defined distribution $p_{\mathrm{ref}}\, e^{r/\beta} / Z$, the fixed point of the Fisher--Rao flow of $-\mathbb{E}_P[r] + \beta\, \mathrm{KL}(P \,\|\, p_{\mathrm{ref}})$ (\cref{app:bg_fisher_rao}), so each generated sample is reweighted by $A_i$ and none is moved.
Our c-transform term also needs no density, and it moves each sample toward nearby group members of higher advantage instead of reweighting it.

\subsection{Optimal-transport views of policy updates}

\paragraph{Reward-guided fine-tuning via Wasserstein gradient flow.}
Hwang et al.~\cite{hwang2026reward} fine-tune a one-step generator toward the reward-tilted target $\nu_r = e^{\beta r} \nu / Z$, where $\nu$ is the distribution of the pretrained model, $r$ is the reward, $\beta > 0$ is an inverse temperature and $Z$ is the normalizer.
They discretize the Wasserstein gradient flow of an $f$-divergence to $\nu_r$ by the JKO scheme:
\begin{align}
\rho_{k+1} = \arg\min_{\rho \in \mathcal{P}_2(\mathcal{X})}\; \frac{1}{2\tau} W_2^2(\rho_k, \rho) + D_f(\rho \,\|\, \nu_r), \nonumber
\end{align}
where $\rho_k$ is the distribution of the generator at iteration $k$, $\tau > 0$ is the step size and $D_f$ is an $f$-divergence.
For the KL divergence,
\begin{align}
\mathrm{KL}(\rho \,\|\, \nu_r) = -\beta\, \mathbb{E}_\rho[r] + \mathrm{KL}(\rho \,\|\, \nu) + \log Z, \nonumber
\end{align}
so the functional is the expected reward plus a KL regularizer toward the pretrained model, and its minimizer is $\nu_r$.
Each step is solved through its semi-dual unbalanced-OT form: a potential network and a transport map are trained adversarially, and the integral against $\nu_r$ is estimated with importance weights $e^{\beta r}$ on samples of the current model.

\paragraph{Relation to and difference from JKO-GRPO.}
Compared with Hwang et al., our functional is $F(\rho) = -\int A\, \mathrm{d}\rho$, which is linear in $\rho$ and has no divergence term.
This makes three differences.
First, because $F$ is linear, the JKO step decouples over points and its solution is the c-transform map (Lemma~\ref{lem:jko_linear}); restricted to the sampled group it is computed in closed form (Lemma~\ref{lem:m_group_jko}), so our step needs no potential network, no adversarial training and no density of $\rho$, whereas the divergence $D_f(\rho \,\|\, \nu_r)$ depends on the density of $\rho$, which is why their step needs a dual potential.
Second, the closeness of the new policy to the old one is controlled in our step only by the transport cost $W_2^2/(2\tau)$, while their step adds a divergence toward a fixed target.
Third, their setting is single-decision generation; ours is the advantage functional of a sequential MDP, applied to the action chunks of a multi-step flow-matching VLA policy and combined with the likelihood-ratio term of Flow-GRPO.
In the inner denoising MDP, JKO-GRPO keeps the clipped likelihood-ratio loss of Flow-GRPO on the Gaussian transitions of the hybrid sampler, without the KL penalty toward the reference model ($\beta = 0$): the closeness of the new policy to the old one is controlled by the transport cost of the c-transform term and by the clipping of the ratio.

We do not include this method as a baseline because its code is not publicly available.

\paragraph{Wasserstein policy optimization.}
Zhang et al.~\cite{zhang2018policy} cast policy optimization as a Wasserstein gradient flow discretized by the JKO scheme.
WPO~\cite{pfau2025wasserstein} starts from the Wasserstein gradient flow of the expected action value, $\partial_t \pi(\cdot \mid s) = -\nabla_a \cdot \big(\pi(\cdot \mid s)\, \nabla_a Q^\pi(s, \cdot)\big)$, in which every action moves along $\nabla_a Q^\pi$, and projects it onto the policy parameters $\theta$ by choosing the parameter change whose induced density change is closest in KL divergence to that of the flow.
This gives the update
\begin{align}
\theta_{t+1} &= \theta_t + F(\theta)^{-1}\, \mathbb{E}_{s,\, a \sim \pi_\theta}\Big[\nabla_\theta \nabla_a \log \pi_\theta(a \mid s)\, \nabla_a Q^\pi(s, a)\Big], \nonumber \\
F(\theta) &= \mathbb{E}_{s,\, a \sim \pi_\theta}\Big[\nabla_\theta \log \pi_\theta(a \mid s)\, \nabla_\theta \log \pi_\theta(a \mid s)^\top\Big], \nonumber
\end{align}
where $Q^\pi$ is the learned action-value function, $\nabla_\theta \nabla_a \log \pi_\theta(a \mid s)$ is the matrix of mixed derivatives of the policy log-density (parameters by action dimensions), and $F(\theta)$ is the Fisher information matrix of the policy.
Like deterministic policy gradients, the update uses $\nabla_a Q^\pi(s, a)$ at the sampled actions, and like classic policy gradients it applies to stochastic policies with arbitrary action distributions; it requires a critic differentiable in the action and the policy log-density.
Our step follows the same transport direction without $\nabla_a Q$ and without a policy density: the displacement $(y_i - x_i)/\tau$ toward the c-transform target is a critic-free estimate built from group advantages and pairwise distances (\cref{app:m_soft}), and it is projected onto the flow sampler by regression through the sampler with the stored noise.


\section{Controlled 2D circle study}
\label{sec:supp-circle}

We first evaluate transport-augmented policy optimization in a
controlled 2D task without branching.
We then present implementation details and additional evaluations
on the robotic VLA benchmarks.

\subsection{Task and initialization}

The observation is a goal $g(\phi)=2(\cos\phi,\sin\phi)$,
with an independent uniform training angle per group.
A complete trajectory generates $x\in\mathbb{R}^2$,
with reward $R(x,g)=-\|x-g\|_2^2$ and success
$\|x-g\|_2<0.2$.
Success depends on the assigned goal, not merely membership
in the radial band $1.8<\|x\|_2<2.2$.
No branching is used.

Each starting model is pretrained on a standard Gaussian
target, without circle goals or successful demonstrations.
For independent $\epsilon,X\sim\mathcal{N}(0,I_2)$ and
$t\sim\mathcal{U}[0,1]$,
\begin{equation}
\begin{aligned}
 z_t&=(1-t)\epsilon+tX,\qquad u=X-\epsilon,\qquad\mathcal{L}_{\mathrm{FM}}
 =\tfrac12\mathbb{E}\|v_\theta(z_t,t)-u\|_2^2.
\end{aligned}
\label{eq:supp-circle-fm}
\end{equation}
The MLP has three width-128 SiLU hidden layers and two outputs.
Pretraining uses 2,000 Adam updates, batch size 1,024 and
initial learning rate $10^{-3}$ with cosine annealing.
Two zero-initialized, trainable goal-input columns are appended
to the first layer, preserving its initial output;
all weights are optimized during RL.

\subsection{Policy optimization and transport}

For $T=10$, $h=1/T$, $t_k=k/T$, $a=0.8$ and
$v_k=v_\theta(z_k,t_k,g)$, the implemented sampler is
\begin{equation}
\begin{aligned}
\mu_{\theta,k}
&=z_k+h\left[v_k-\frac{a^2}{2}(z_k-t_kv_k)\right],\\
z_{k+1}
&=\mu_{\theta,k}+s_k\epsilon_{k+1},\\
s_k
&=a\sqrt{(1-t_k)h},\qquad z_0=\epsilon_0.
\end{aligned}
\label{eq:supp-circle-sde}
\end{equation}
The complete noise tape $\epsilon_{0:10}$ contains 11 independent standard Gaussian two-vectors; $x=z_{10}$. All 10 transitions are stochastic.

Each round collects four same-goal groups of $G=20$ paths under the old policy.
Within each group,
$A_i=(R_i-\bar R)/(s_R+10^{-8})$,
where $s_R$ uses denominator $G-1$.
The terminal advantage applies to every transition of its path. For saved transitions, Gaussian coordinate log densities are summed, giving
\begin{equation}
\begin{aligned}
 \ell_{\theta,i,k}
 &=-\frac{\|z_{i,k+1}
       -\mu_{\theta,k}(z_{i,k},g_i)\|_2^2}{2s_k^2},\\
 q_{i,k}
 &=\exp(\ell_{\theta,i,k}-\ell_{\mathrm{old},i,k}).
\end{aligned}
\label{eq:supp-circle-ratio}
\end{equation}
Policy-independent Gaussian constants cancel.
Ratios are per transition, not products over a path.
For minibatches of $B=20$ paths, the shared clipped loss is
\begin{equation}
\begin{aligned}
 \mathcal{L}_{\mathrm{clip}}
 =-\frac1{BT}\sum_{i,k}
 \min\big\{&q_{i,k}A_i,
 \operatorname{clip}
 (q_{i,k},1-\eta,1+\eta)A_i\big\}.
\end{aligned}
\label{eq:supp-circle-clip}
\end{equation}
We write the corresponding Flow-GRPO objective as
$\mathcal{J}_{\mathrm{FG}}=-\mathcal{L}_{\mathrm{clip}}$.
Saved rollout data, old likelihoods and advantages remain
detached during the round.

JKO-GRPO constructs weights $p_{ij}$ and targets $y_i$
using \eqref{eq:m_weights} and \eqref{eq:m_target},
restricted to each original same-goal group of rewarded
SDE endpoints.
Targets are fixed without gradients for the round.
The current policy replays the same complete noise tape
and goal, differentiating through all ten steps:
\begin{equation}
\begin{aligned}
 m_i&=\mathbb{1}[\max_j A_j-\min_j A_j>10^{-8}]\,
       \mathbb{1}[p_{ii}<0.5],\\
 \mathcal{L}_{\mathrm{ct}}
 &=\frac1B\sum_i m_i
 \|f_\theta(\epsilon_{i,0:10};g_i)
   -\mathrm{sg}(y_i)\|_2^2,\\
 \mathcal{L}
 &=\lambda_1\mathcal{L}_{\mathrm{ct}}
   -\lambda_2\mathcal{J}_{\mathrm{FG}}.
\end{aligned}
\label{eq:supp-circle-loss}
\end{equation}
We fix $\lambda_1{=}5$ and $\lambda_2{=}1$ in the circle study.
Coordinate errors are summed and the denominator includes
inactive rows.
Reward statistics and transport distances/weights use
float64; policy computation and targets use float32.
This study fits rewarded SDE endpoints using SDE replay of the
full noise tape.
The robotic VLA experiments use deterministic ODE replay by default
(\cref{app:m_impl}); an SDE replay comparison is reported in
\cref{tab:supp-replay}.

\begin{table}[tbp]
\centering
\small
\begin{tabular}{@{}lc@{}}
\toprule
Setting & Value \\
\midrule
Flow steps / noise scale & $10/0.8$ \\
Goal groups / paths per group & $4/20$ \\
Collection rounds / paths per round & $200/80$ \\
Optimization passes / minibatches per pass & $2/4$ \\
Paths per minibatch & $20$ \\
RL Adam learning rate / gradient-norm limit & $10^{-3}/1$ \\
PPO clipping $\eta$ / KL coefficient & $0.3/0$ \\
JKO-GRPO: fixed $\lambda_1$ / $\tau$ / $\kappa$ & $5/1/0.05$ \\
\bottomrule
\end{tabular}
\caption{Circle-study hyperparameters.
Shared settings apply to both methods;
the final row applies only to JKO-GRPO, with $\lambda_2{=}1$ fixed.}
\label{tab:supp-circle-hparams}
\end{table}

Groups and targets are defined before minibatch shuffling.
The protocol gives 16,000 rewarded paths,
800 training goal contexts and 1,600 Adam updates per run.
All settings, including the hyperparameters, are fixed in this execution.

\subsection{Evaluation protocol and metrics}

We evaluate all 9 combinations of 3 pretraining seeds
and 3 RL seeds, using the same seed combinations across methods.
Each pair shares its initial checkpoint, noise/goal tapes
and shuffle stream; a given RL tape is also shared across bases.
Thus the grid contains 3 pretrained models,
not nine independent initializations.
A disposable startup-check base is excluded.

Circle evaluations use the stochastic sampler on fixed held-out
noise and goal tapes.
Curves evaluate 64 goals with 20 paths each (1,280 outputs)
at rounds 0--200 on fixed held-out inputs.
Final evaluation uses a separate tape of 256 goals and
20 paths each (5,120 outputs).
Angles are uniformly stratified with independent within-bin
jitter.
Evaluation never contributes to training or its sample budget;
all nine settings receive equal weight.

For held-out reward or success $m$, normalized AUC is
\begin{equation}
\begin{aligned}
 \operatorname{AUC}_{N}(m)
 &=\frac1N\int_0^N\bar m(n)\,\mathrm{d}n,\\
 \operatorname{AUC}_{t}(m)
 &=\frac1{t_*}\int_0^{t_*}\bar m(t)\,\mathrm{d}t.
\end{aligned}
\label{eq:supp-circle-auc}
\end{equation}
Here $N=16{,}000$ and $t_*=3.84$\,s is the shortest
final measured time among all 18 runs.
We average over settings, using trapezoidal integration and
per-run linear interpolation without extrapolation.
Numerical time AUCs use these fresh measured clocks;
the figure's fixed reference-clock panels are identified
separately below.
Descriptive 95\% intervals use 10,000 crossed-bootstrap
replicates, resampling base rows and RL columns independently
while preserving method pairing.

\subsection{Results and computational cost}

JKO-GRPO improves reward/\allowbreak sample AUC in all nine settings
and final success in seven.
The paired mean gains are $0.068$
(descriptive 95\% interval $[0.056,0.079]$)
and $13.73$ percentage points ($[2.82,27.32]$), respectively.
Reward/\allowbreak time AUC decreases by $0.022$,
with difference interval $[-0.047,\allowbreak -0.003]$.
Success AUCs are secondary post-run descriptive summaries.

\begin{figure}[tbp]
\centering
\includegraphics[width=0.8\linewidth]
{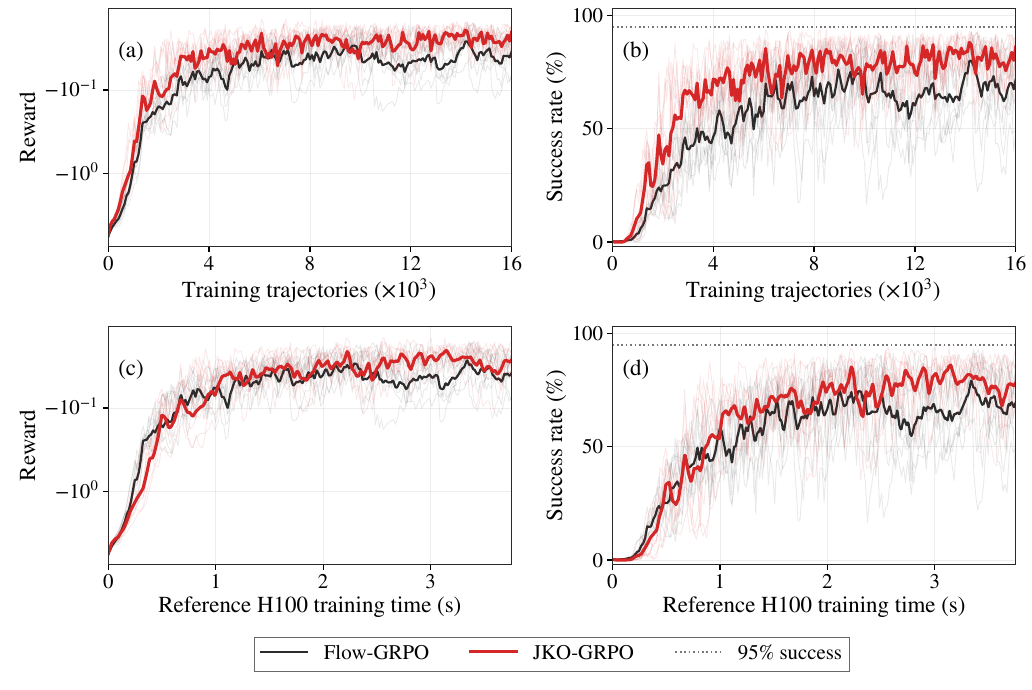}
\caption{\textbf{Circle learning curves.}
Top: held-out reward and assigned-goal success versus
rewarded training trajectories.
Bottom: the same evaluations placed at fixed, previously
measured H100 reference timestamps, matched by method and
seed pair; common plotted support is
$0$--$3.75$\,s.
Thin curves show all nine settings;
thick black/red curves show Flow-GRPO/JKO-GRPO means.
The numerical time AUCs in \cref{tab:supp-circle-results}
instead use the fresh measured clocks over
$0$--$3.84$\,s.}
\label{fig:supp-circle-learning-curves}
\end{figure}

\begin{table}[tbp]
\centering
\small
\begin{tabular}{@{}lrr@{}}
\toprule
Metric & Flow-GRPO & JKO-GRPO \\
\midrule
Reward/sample AUC & $-0.255$ & $\mathbf{-0.187}$ \\
Success/sample AUC (\%) & $54.32$ & $\mathbf{67.88}$ \\
Reward/time AUC & $\mathbf{-0.265}$ & $-0.288$ \\
Success/time AUC (\%) & $53.81$ & $\mathbf{60.41}$ \\
Final assigned-goal success (\%) & $72.05$ & $\mathbf{85.77}$ \\
Final mean squared distance & $0.035$ & $\mathbf{0.020}$ \\
Training-loop time (s) & $\mathbf{3.961}$ & $6.088$ \\
\bottomrule
\end{tabular}
\caption{Nine-setting means.
Higher reward/success AUC and final success are better;
lower squared distance and training time are better.
Success AUCs are expressed as percentages and remain
distinct from final success.}
\label{tab:supp-circle-results}
\end{table}

At 16,000 paths, JKO-GRPO takes $1.54\times$ the H100
training-loop time, adding transport-target construction
and differentiable replay.
Both methods use the same compilation settings with shape
padding disabled.
Synchronized timing includes collection, reward/statistics
and optimization, including JKO target/replay computation;
it excludes pretraining, input preparation,
compilation/warmup, evaluation, I/O and queue waiting.
These are training-loop measurements on one H100,
not end-to-end GPU-hour savings.

\section{Robotic VLA experiments}
\label{sec:supp-robot}

This section provides implementation details, evaluation protocols,
and additional results for the robotic VLA benchmarks.

\subsection{Implementation details}
\label{supp:impl}

\paragraph{Hyperparameters.}
\cref{tab:supp-hparams} lists all settings.
Within a benchmark, every method starts from the same SFT
checkpoint and uses the same environment, trajectory budget,
group size, learning rate and number of optimizer steps per iteration.
The per-benchmark values (chunk length, denoising steps,
rollout noise, episode length, learning rate) are those released by
$\pi_{\texttt{RL}}$~\cite{pirl2026}.
Flow-GRPO is recovered by setting $\lambda_1{=}0$,
$\lambda_2{=}1$, and $M{=}0$ in JKO-GRPO.

\begin{table*}[t]
\centering
\small
\setlength{\tabcolsep}{5pt}
\begin{tabular}{@{}l|c|cccc|c@{}}
\toprule
 & ManiSkill & \multicolumn{4}{c|}{LIBERO} & MetaWorld \\
 & & Spatial & Object & Goal & Long & MT50 \\
\midrule
action dimension & 7 & 7 & 7 & 7 & 7 & 4 \\
chunk length & 5 & 5 & 5 & 5 & 10 & 5 \\
denoising steps & 4 & 3 & 5 & 5 & 4 & 5 \\
rollout noise level (Flow-SDE) & 0.5 & 0.5 & 0.3 & 0.3 & 0.5 & 0.5 \\
max.\ episode steps & 80 & 240 & 240 & 320 & 480 & 100 \\
\midrule
trajectories per iteration & \multicolumn{6}{c}{512 ($G{=}8$ per group)} \\
parallel envs $\times$ rollout rounds & 256 $\times$ 2 & \multicolumn{5}{c}{128 $\times$ 4} \\
chunks per optimizer step & \multicolumn{6}{c}{2048} \\
update epochs & 1 & 1 & 1 & 1 & 4 & 1 \\
learning rate & $7.91{\times}10^{-6}$ & \multicolumn{5}{c}{$5{\times}10^{-6}$} \\
learning-rate schedule & \multicolumn{5}{c|}{constant} & cosine \\
PPO clip $\epsilon$ / dual clip & \multicolumn{6}{c}{0.2 / 3.0} \\
gradient-norm clip & \multicolumn{6}{c}{1.0} \\
group filter (success rate) & \multicolumn{6}{c}{$[0.1, 0.9]$} \\
iterations & 400 & 40 & 40 & 40 & 40 & 40 \\
\midrule
transport weight $\lambda_1$ & 10 & 1200 & 1200 & 1200 & 1200 & 0.8 \\
cost scale $\tau$ & 1 & 0.3 & 0.1 & 0.1 & 0.1 & 12 \\
softmax temperature $\kappa$ & \multicolumn{6}{c}{0.05} \\
branch points $M$ & \multicolumn{6}{c}{4} \\
\bottomrule
\end{tabular}
\caption{\textbf{Hyperparameters.}
The upper and middle blocks are shared by all methods within a
benchmark; the lower block is specific to JKO-GRPO,
with $\lambda_2{=}1$ throughout.
The group filter, as in $\pi_{\texttt{RL}}$, drops groups whose
success rate lies outside the interval.}
\label{tab:supp-hparams}
\end{table*}

\paragraph{Branch points.}
\label{supp:branch}
With $M$ branch points, $M$ groups of $G$ environments share one initial state. One environment of the first group runs an unmodified source trajectory from $\pi_{\theta_{\mathrm{old}}}$. Group $m$ draws its own chunk index $t_m$, with the $M$ indices drawn without replacement from the chunks of an episode. At the boundary before chunk $t_m$, the simulator state of the source is copied into the environments of group $m$, which then sample their own continuations. If the source has already terminated, the group keeps its trajectories from the initial state ($t_m{=}0$). Rewards before $t_m$ are discarded, so advantages are computed on the post-branch segment, and chunks before $t_m$ are masked out of both losses. The transport target acts on chunk $t_m$, where the $G$ samples share a state. Under a fixed budget of 512 trajectories, $M{=}0$, 1, 2 and 4 use 64, 64, 32 and 16 initial states per iteration.

\paragraph{DGPO.}
\label{supp:dgpo}
We follow the official implementation of DGPO~\cite{luo2026dgpo}: deterministic ODE rollouts with the denoising steps of \cref{tab:supp-hparams}, forward-process noise shared within a group, timesteps drawn from $[0.3,1]$, advantages clipped at 5, a KL term to the frozen SFT policy (weight 0.04), and PPO-style clipping of the denoising loss ($10^{-3}$). The one change is the temperature $\beta$, which scales the group's denoising-loss differences inside the sigmoid and does not scale the gradient. In our action space, these differences are small enough that the default $\beta{=}100$ keeps the sigmoid weight at $0.5$ for every group; we use $\beta{=}10^{4}$. The number of optimizer steps per iteration is matched to Flow-GRPO.

\subsection{Evaluation protocol}
\label{supp:eval}

\paragraph{Evaluation states and metrics.}
All robotic evaluations use the deterministic sampler,
and an episode counts as a success if the task is solved at any step.
On ManiSkill, following $\pi_{\texttt{RL}}$, we evaluate on 2048 layouts drawn from the training layout pool with the default environment seed, which is also used in training, so these layouts can occur during training. Re-evaluating the SFT policy and the iteration-400 checkpoints on 2048 layouts that never occur in training changes every success rate by at most 1.1 points and leaves the ranking unchanged (SFT 55.8, Flow-GRPO 84.1, DGPO 57.9, JKO-GRPO 90.6).
For generalization, we follow RL4VLA~\cite{liu2025rl4vla}
and evaluate the iteration-400 checkpoints on 13 unseen scenes,
512 episodes each.
LIBERO and MetaWorld follow the $\pi_{\texttt{RL}}$ protocol:
training and evaluation use the same 500 initial states per suite,
and there is no held-out split.
On LIBERO and MetaWorld, we evaluate every 10 iterations.

\paragraph{Checkpoint and hyperparameter selection.}
On LIBERO and MetaWorld, we evaluated four to eight
$(\lambda_1,\tau)$ configurations per benchmark,
fixing $\lambda_2{=}1$.
Some runs were stopped early; final configurations were selected
by success at iteration 40 among runs reaching that checkpoint.
The robotic benchmark results are from a single training run
per configuration.

\subsection{Expanded benchmark results}\label{supp:results}

\paragraph{ManiSkill every 40 iterations.}
\cref{tab:supp-ms-iters} lists every evaluation of the ManiSkill runs, including the branch-point ablation of the main paper.

\begin{table*}[t]
\centering
\small
\setlength{\tabcolsep}{4.5pt}
\begin{tabular}{@{}l|cccccccccc@{}}
\toprule
Iteration & 40 & 80 & 120 & 160 & 200 & 240 & 280 & 320 & 360 & 400 \\
\midrule
SFT & \multicolumn{10}{c}{56.2} \\
\midrule
Flow-GRPO & 74.5 & 77.1 & 78.4 & 79.0 & 79.5 & 80.5 & 81.9 & 82.7 & 81.2 & 84.7 \\
DGPO & 60.5 & 59.0 & 61.7 & 61.4 & 61.3 & 61.6 & 59.0 & 60.7 & 58.2 & 58.1 \\
\midrule
JKO-GRPO, $M{=}0$ & \textbf{78.4} & \textbf{82.3} & 83.2 & 83.6 & 86.4 & 87.8 & 86.7 & 86.5 & 88.0 & 89.2 \\
JKO-GRPO, $M{=}1$ & 74.8 & 80.4 & 84.3 & 82.7 & 83.4 & 84.9 & 82.4 & 84.8 & 81.3 & 85.5 \\
JKO-GRPO, $M{=}2$ & 75.2 & 80.8 & \textbf{87.0} & 84.6 & 86.6 & 87.2 & 86.0 & 88.0 & 88.9 & 87.6 \\
JKO-GRPO, $M{=}4$ & 75.6 & 77.2 & 83.4 & \textbf{87.3} & \textbf{89.6} & \textbf{90.4} & \textbf{89.8} & \textbf{88.3} & \textbf{89.5} & \textbf{90.3} \\
\bottomrule
\end{tabular}
\caption{\textbf{ManiSkill every 40 iterations.} Success rate (\%) on the 2048 evaluation layouts, extending \cref{tab:main} of the main paper. All RL methods use 512 trajectories per iteration. $M$ is the number of branch points ($M{=}0$: groups at the initial state only); $M{=}4$ is the default and the JKO-GRPO row of \cref{tab:main}. The best value in each column is in bold.}
\label{tab:supp-ms-iters}
\end{table*}

\paragraph{LIBERO and MetaWorld every 10 iterations.}
\cref{tab:supp-iters} extends the main LIBERO and MetaWorld table to every evaluation. At iteration 40, JKO-GRPO achieves the highest success rate on four of the five suites and ranks second on LIBERO-Long, within 0.8 percentage points of Flow-GRPO (60.2\% versus 61.0\%).

\begin{table}[t]
\centering
\small
\setlength{\tabcolsep}{4pt}
\begin{tabular}{@{}ll|cccc@{}}
\toprule
 & Method & 10 & 20 & 30 & 40 \\
\midrule
\multirow{3}{*}{Spatial}
 & Flow-GRPO & \textbf{92.8} & \textbf{92.0} & \textbf{92.8} & 90.2$^{*}$ \\
 & DGPO & 81.4 & 81.0 & 81.8 & 80.8 \\
 & JKO-GRPO & 90.2$^{*}$ & 91.0$^{*}$ & 92.0$^{*}$ & \textbf{92.0} \\
\midrule
\multirow{3}{*}{Object}
 & Flow-GRPO & \textbf{99.8} & \textbf{99.6} & \textbf{99.2} & 96.4$^{*}$ \\
 & DGPO & 94.6 & 91.0 & 94.4$^{*}$ & 94.8 \\
 & JKO-GRPO & 99.6$^{*}$ & 99.2$^{*}$ & \textbf{99.2} & \textbf{98.0} \\
\midrule
\multirow{3}{*}{Goal}
 & Flow-GRPO & \textbf{84.8} & 84.2 & \textbf{89.4} & 87.4$^{*}$ \\
 & DGPO & 84.6$^{*}$ & 86.0$^{*}$ & 84.2$^{*}$ & 85.6 \\
 & JKO-GRPO & 82.2 & \textbf{86.6} & 83.0 & \textbf{87.8} \\
\midrule
\multirow{3}{*}{Long}
 & Flow-GRPO & \textbf{58.2} & \textbf{54.0} & \textbf{57.8} & \textbf{61.0} \\
 & DGPO & 41.8 & 40.4 & 45.2 & 43.8 \\
 & JKO-GRPO & 51.2$^{*}$ & 49.6$^{*}$ & 54.4$^{*}$ & 60.2$^{*}$ \\
\midrule
\multirow{3}{*}{MT50}
 & Flow-GRPO & \textbf{58.6} & 57.8$^{*}$ & 59.2$^{*}$ & 61.0$^{*}$ \\
 & DGPO & 52.6 & 48.0 & 48.8 & 51.4 \\
 & JKO-GRPO & 58.4$^{*}$ & \textbf{61.2} & \textbf{60.4} & \textbf{62.0} \\
\bottomrule
\end{tabular}
\caption{\textbf{LIBERO and MetaWorld every 10 iterations.}
Success rate (\%) on the 500 initial states per suite.
The SFT policy achieves 82.2, 96.4, 86.4, and 45.0 on
LIBERO-Spatial, Object, Goal, and Long, respectively,
and 47.0 on MetaWorld.
Bold and $^{*}$ indicate the best and second-highest distinct
scores, respectively, among the three RL methods within each
suite and iteration.}
\label{tab:supp-iters}
\end{table}

\paragraph{Generalization per scene.}
\cref{tab:supp-ood} lists the success rate in each of the
13 unseen scenes summarized in the main paper.

\begin{table}[t]
\centering
\small
\setlength{\tabcolsep}{3pt}
\begin{tabular}{@{}ll|cccc@{}}
\toprule
 & Scene & SFT & Flow-GRPO & DGPO & JKO-GRPO \\
\midrule
\multirow{5}{*}{\rotatebox[origin=c]{90}{Vision}}
 & Vis-Img & 60.0 & 85.0 & 64.6 & \textbf{88.1} \\
 & Tex-0.3 & 44.9 & 78.9 & 47.7 & \textbf{87.7} \\
 & Tex-0.5 & 40.8 & 68.4 & 44.9 & \textbf{77.0} \\
 & Whole-0.3 & 45.9 & 80.9 & 45.9 & \textbf{86.3} \\
 & Whole-0.5 & 35.5 & 67.4 & 39.8 & \textbf{74.2} \\
 & \emph{Mean} & 45.4 & 76.1 & 48.6 & 82.7 \\
\midrule
\multirow{5}{*}{\rotatebox[origin=c]{90}{Semantic}}
 & Instruct & 57.6 & 84.2 & 59.0 & \textbf{92.8} \\
 & Obj & 39.5 & 67.8 & 41.8 & \textbf{78.1} \\
 & Plate & 52.7 & 82.2 & 55.1 & \textbf{89.1} \\
 & M-Obj & 19.3 & 33.4 & 18.8 & \textbf{46.3} \\
 & M-Plate & 14.5 & 19.1 & 17.0 & \textbf{30.9} \\
 & \emph{Mean} & 36.7 & 57.3 & 38.3 & 67.4 \\
\midrule
\multirow{3}{*}{\rotatebox[origin=c]{90}{Exec.}}
 & Pos-Chg & 19.9 & 31.8 & 15.4 & \textbf{34.2} \\
 & Pos & 39.1 & 53.1 & 35.9 & \textbf{66.2} \\
 & EE-Pose & 30.5 & 52.5 & 27.1 & \textbf{64.6} \\
 & \emph{Mean} & 29.8 & 45.8 & 26.2 & 55.0 \\
\midrule
\multicolumn{2}{@{}l|}{All 13 scenes} & 38.5 & 61.9 & 39.5 & 70.4 \\
\bottomrule
\end{tabular}
\caption{\textbf{Generalization per scene}
(ManiSkill, iteration 400).
Success rate (\%) in each of the 13 unseen scenes of
RL4VLA~\cite{liu2025rl4vla}, 512 episodes each.
Scene names as in the main-paper figure.}
\label{tab:supp-ood}
\end{table}

\subsection{Failure analysis and qualitative rollouts}

\paragraph{Failure modes.}
\cref{tab:supp-fail} splits the failed ManiSkill episodes
at iteration 200.
A grasp is stable if the object is held for 5 consecutive steps.
These numbers come from a separate evaluation of the same
checkpoints on the same 2048 layouts, with failure logging enabled;
success is 80.4\% for Flow-GRPO (79.5\% in the main evaluation)
and 89.6\% for JKO-GRPO (identical).
Most of the gain is in the first mode:
Flow-GRPO often moves the empty gripper to the plate.

\begin{table}[t]
\centering
\small
\setlength{\tabcolsep}{3pt}
\begin{tabular}{@{}l|cc@{}}
\toprule
Failure mode & Flow-GRPO & JKO-GRPO \\
\midrule
No grasp, gripper at plate & 11.3 & \textbf{3.5} \\
No grasp, other & 2.7 & \textbf{2.3} \\
Grasp, over plate, not placed & 1.8 & \textbf{0.9} \\
Grasp, not brought to plate & 3.8 & 3.8 \\
\midrule
Total failures & 19.6 & \textbf{10.4} \\
\bottomrule
\end{tabular}
\caption{\textbf{Failure modes}
(ManiSkill, iteration 200), as \% of the 2048 episodes.
The gripper is at the plate if it comes within 8\,cm of the
plate centre; the object is over the plate if it comes within
the plate's half-size plus 2\,cm.}
\label{tab:supp-fail}
\end{table}

\begin{figure*}[t]
\centering
\includegraphics[width=0.49\linewidth]{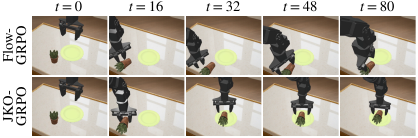}\hfill
\includegraphics[width=0.49\linewidth]{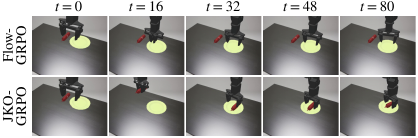}
\caption{\textbf{Rollouts from two initial states}
(ManiSkill, iteration 200, deterministic sampler).
Each pair of rows starts from the same layout;
$t$ is the environment step.}
\label{fig:supp-qual}
\end{figure*}

\paragraph{Qualitative rollouts.}
\cref{fig:supp-qual} shows two more ManiSkill layouts
in which the two policies differ.
Flow-GRPO tips the pot over or reaches the plate without the object;
JKO-GRPO grasps the object and places it.

\subsection{Additional ablations}

\paragraph{Regression targets.}
\cref{tab:supp-raft} extends the regression-target ablation
of the main paper.
Without the GRPO loss, RAFT~\cite{dong2023raft} trains on
successful chunks alone and achieves only 4.8\% success
at iteration 80, recovering to 64.3\% at iteration 160
but remaining below Flow-GRPO.
Adding a RAFT-style target to Flow-GRPO improves early learning
in both budget settings, but its largest gain at iteration 160
is 2.4 percentage points.
The transport target achieves the highest success rate
at both checkpoints under both budgets.

\begin{table}[t]
\centering
\small
\setlength{\tabcolsep}{4pt}
\begin{tabular}{@{}l|cc@{}}
\toprule
Regression target & Iter.\ 80 & Iter.\ 160 \\
\midrule
\multicolumn{3}{@{}l}{\emph{320 trajectories per iteration}} \\
None (Flow-GRPO) & 72.4 & 79.5 \\
RAFT only, no GRPO loss & 4.8 & 64.3 \\
RAFT-style target, first chunk & 78.9$^{*}$ & 79.2 \\
RAFT-style target, all chunks & 78.8 & 80.9$^{*}$ \\
Transport target & \textbf{81.8} & \textbf{81.0} \\
\midrule
\multicolumn{3}{@{}l}{\emph{512 trajectories per iteration}} \\
None (Flow-GRPO) & 77.1 & 79.0 \\
RAFT-style target, first chunk & 79.9$^{*}$ & 81.4$^{*}$ \\
Transport target & \textbf{82.3} & \textbf{83.6} \\
\bottomrule
\end{tabular}
\caption{\textbf{Regression targets on ManiSkill} ($M{=}0$).
The transport target uses $\lambda_1{=}10$,
$\lambda_2{=}1$, and $\tau{=}1$.
RAFT-style targets use a flow-matching loss on executed chunks
from successful rollouts, with weight 5 for all-chunk regression
and RAFT only, and 80 for first-chunk regression.
Within each budget and checkpoint, best scores are bolded
and second-best scores are marked with $^{*}$.}
\label{tab:supp-raft}
\end{table}

\paragraph{ODE or SDE replay.}
Rollouts use the Flow-SDE sampler of $\pi_{\texttt{RL}}$,
in which one denoising step is stochastic, while our transport
loss replays the deterministic ODE from the stored initial noise.
We also tried replaying the executed SDE chain:
at the stochastic step, the replay uses the current policy's
mean plus the stored noise of that step, so that at
$\theta{=}\theta_{\mathrm{old}}$ it reproduces the executed action.
\cref{tab:supp-replay} shows that the SDE replay is not better.
We use ODE replay for the robotic VLA experiments.

\begin{table}[t]
\centering
\small
\setlength{\tabcolsep}{4pt}
\begin{tabular}{@{}ll|cccc@{}}
\toprule
 & Replay & 10 & 20 & 30 & 40 \\
\midrule
Spatial & ODE & 88.6 & 91.6 & 93.8 & 91.2 \\
$\lambda_1{=}2000$, $\tau{=}0.005$ & ODE+SDE & 89.4 & 90.4 & 89.8 & 91.2 \\
\midrule
Long & ODE & 51.2 & 49.6 & 54.4 & 60.2 \\
$\lambda_1{=}1200$, $\tau{=}0.1$ & ODE+SDE & 50.8 & 52.6 & 49.4 & 51.8 \\
\bottomrule
\end{tabular}
\caption{\textbf{ODE vs.\ SDE replay} on LIBERO.
Success rate (\%) on the 500 initial states at iterations 10--40,
with $\lambda_2{=}1$ fixed.}
\label{tab:supp-replay}
\end{table}

\subsection{Transport scaling and sensitivity}

\paragraph{Loss scales.}
Since action representations, chunk lengths, and transport-loss scales
differ across benchmarks, we use benchmark-specific $(\lambda_1,\tau)$,
while fixing $\lambda_2{=}1$ and $\kappa{=}0.05$ throughout.
\cref{tab:supp-scale} summarizes the selected hyperparameters and
the resulting unweighted and weighted transport losses.

\begin{table}[t]
\centering
\small
\setlength{\tabcolsep}{3pt}
\begin{tabular}{@{}l|rr|ccc@{}}
\toprule
Benchmark
 & $\lambda_1$
 & $\tau$
 & $\overline{\mathcal L_{\mathrm{ct}}}$
 & $\overline{\lambda_1\mathcal L_{\mathrm{ct}}}$
 & $r_{\mathrm{loss}}$ \\
\midrule
ManiSkill
 & 10 & 1
 & $7.5{\times}10^{-4}$
 & $7.5{\times}10^{-3}$
 & 23 \\
\midrule
LIBERO-Spatial
 & 1200 & 0.3
 & $4.6{\times}10^{-5}$
 & $5.5{\times}10^{-2}$
 & 12.5 \\
LIBERO-Object
 & 1200 & 0.1
 & $3.5{\times}10^{-6}$
 & $4.3{\times}10^{-3}$
 & 3.0 \\
LIBERO-Goal
 & 1200 & 0.1
 & $9.3{\times}10^{-6}$
 & $1.1{\times}10^{-2}$
 & 11.3 \\
LIBERO-Long
 & 1200 & 0.1
 & $2.6{\times}10^{-5}$
 & $3.1{\times}10^{-2}$
 & 4.0 \\
\midrule
MetaWorld MT50
 & 0.8 & 12
 & $1.3{\times}10^{-3}$
 & $1.1{\times}10^{-3}$
 & 0.3 \\
\bottomrule
\end{tabular}
\caption{\textbf{Loss scales under the reported JKO-GRPO configurations.}
Means are computed from the logged, micro-batch-averaged losses
over 400 training iterations on ManiSkill and 40 on the other benchmarks.
The relative loss scale is
$r_{\mathrm{loss}}
=\overline{\lambda_1\mathcal L_{\mathrm{ct}}}
/\overline{|\lambda_2\mathcal J_{\mathrm{FG}}|}$,
where $\mathcal J_{\mathrm{FG}}$ is the clipped Flow-GRPO objective
in \eqref{eq:total} and $\lambda_2{=}1$ throughout.}
\label{tab:supp-scale}
\end{table}

\paragraph{Hyperparameter sensitivity on ManiSkill.}
We fix $\lambda_2{=}1$ throughout this analysis. \cref{tab:supp-sens} varies $\lambda_1$ and $\tau$ by a factor of three around their default values. All tested configurations outperform Flow-GRPO at both reported checkpoints.

\begin{table}[t]
\centering
\small
\begin{tabular}{@{}cc|cc@{}}
\toprule
$\lambda_1$ & $\tau$ & Iter.\ 80 & Iter.\ 160 \\
\midrule
\multicolumn{2}{c|}{Flow-GRPO} & 72.4 & 79.5 \\
\midrule
10 & 0.3 & 81.0 & 84.3 \\
10 & 1 & 81.8 & 81.0 \\
10 & 3 & 81.7 & \textbf{87.1} \\
3 & 1 & 80.4 & 81.8 \\
30 & 1 & \textbf{82.8} & 80.0 \\
\bottomrule
\end{tabular}
\caption{\textbf{Sensitivity to $\lambda_1$ and $\tau$}
(ManiSkill, $M{=}0$, 320 trajectories per iteration).
Success rate (\%) on the 2048 evaluation layouts.}
\label{tab:supp-sens}
\end{table}

\paragraph{Sensitivity on LIBERO and MetaWorld.}
\cref{tab:supp-sens-other} lists every $(\lambda_1,\tau)$ configuration trained to iteration 40. We report the one with the highest success at iteration 40; on Spatial, $\tau{=}0.1$ falls below Flow-GRPO (87.8\% vs.\ 90.2\%), so $\tau{=}0.3$ is selected.


\begin{table}[t]
\centering
\small
\setlength{\tabcolsep}{3pt}
\begin{tabular}{@{}lrr|cccc@{}}
\toprule
Suite & $\lambda_1$ & $\tau$ & 10 & 20 & 30 & 40 \\
\midrule
\multirow{4}{*}{Spatial}
 & 1200 & 0.1 & 88.6 & 88.4 & 89.0 & 87.8 \\
 & 1200$^\dagger$ & 0.3 & \textbf{90.2} & 91.0 & 92.0 & \textbf{92.0} \\
 & 2000 & 0.005 & 88.6 & \textbf{91.6} & \textbf{93.8} & 91.2 \\
 & 3000 & 0.01 & 86.8 & 90.0 & 92.6 & 90.6 \\
\midrule
\multirow{3}{*}{Object}
 & 1200$^\dagger$ & 0.1 & \textbf{99.6} & \textbf{99.2} & \textbf{99.2} & \textbf{98.0} \\
 & 2000 & 0.005 & 98.8 & 98.0 & 97.8 & 94.8 \\
 & 3000 & 0.01 & 92.2 & 97.2 & 98.2 & \textbf{98.0} \\
\midrule
\multirow{3}{*}{Goal}
 & 1200$^\dagger$ & 0.1 & 82.2 & \textbf{86.6} & 83.0 & \textbf{87.8} \\
 & 2000 & 0.005 & 82.0 & 83.4 & 82.6 & 78.2 \\
 & 3000 & 0.01 & \textbf{86.8} & 85.6 & \textbf{84.0} & 84.8 \\
\midrule
\multirow{2}{*}{Long}
 & 1200$^\dagger$ & 0.1 & \textbf{51.2} & 49.6 & \textbf{54.4} & \textbf{60.2} \\
 & 2000 & 0.005 & 46.8 & \textbf{52.6} & \textbf{54.4} & 57.4 \\
\midrule
\multirow{2}{*}{MT50}
 & 0.8$^\dagger$ & 12 & 58.4 & \textbf{61.2} & \textbf{60.4} & \textbf{62.0} \\
 & 500 & 0.3 & \textbf{60.2} & 59.0 & 58.2 & 58.4 \\
\bottomrule
\end{tabular}
\caption{\textbf{Transport configurations on LIBERO and MetaWorld.}
Success rate (\%) on 500 initial states per suite at iterations 10--40, for every $(\lambda_1,\tau)$ trained to iteration 40.
All use ODE replay, $M{=}4$, and $\lambda_2{=}1$.
$^\dagger$Reported configuration. Best scores within each suite and checkpoint are bolded, including ties.}
\label{tab:supp-sens-other}
\end{table}

\subsection{Computational cost}

Every robotic training run uses one node with four H100 GPUs.
On ManiSkill, an iteration takes 243\,s for Flow-GRPO and
261\,s for JKO-GRPO, about 27 and 29 hours for 400 iterations.
\cref{tab:supp-time} gives the time per iteration on LIBERO
and MetaWorld.
Rollouts and environment steps dominate;
the transport term adds 3--9\% per iteration.

\begin{table*}[t]
\centering
\small
\setlength{\tabcolsep}{4pt}
\begin{tabular}{@{}l|rrr|rrr|r@{}}
\toprule
 & \multicolumn{3}{c|}{time per iteration (s)}
 & \multicolumn{3}{c|}{actor update (s)} & \\
benchmark
 & Flow-GRPO & DGPO & JKO-GRPO
 & Flow-GRPO & DGPO & JKO-GRPO
 & JKO / Flow \\
\midrule
LIBERO-Spatial & 1120 & 1494 & 1148 & 101 & 501 & 131 & 1.03 \\
LIBERO-Object & 851 & 1168 & 899 & 101 & 408 & 152 & 1.06 \\
LIBERO-Goal & 1254 & 1378 & 1317 & 135 & 255 & 202 & 1.05 \\
LIBERO-Long & 1626 & 2092 & 1779 & 397 & 877 & 562 & 1.09 \\
MetaWorld MT50 & 1056 & 1394 & 1114 & 44 & 375 & 65 & 1.06 \\
\bottomrule
\end{tabular}
\caption{\textbf{Time per iteration} on LIBERO and MetaWorld:
median wall-clock time of a training iteration and of its actor
update (s), on four H100 GPUs.
Runs were on different nodes, so the numbers are approximate.}
\label{tab:supp-time}
\end{table*}

\end{document}